\RequirePackage{fix-cm}
\documentclass[twocolumn]{svjour3}          
\smartqed  
\usepackage{graphicx}
\usepackage{mathptmx,epsfig,latexsym,epstopdf,natbib, subfig}

\journalname{}
\begin{document}

\title{A Spiking Neural Learning Classifier System}


\author{Gerard Howard         \and
       Larry Bull \and
	Pier-Luca Lanzi
}


\institute{G. Howard \at
              Faculty of Environment and Technology, University of the West of England,\\
              \email{david4.howard@uwe.ac.uk}           
           \and
          L. Bull \at
               Faculty of Environment and Technology, University of the West of England, 
           \and
           P.-L. Lanzi \at
              Dipartimento di Elettronica e Informazione, Politecnico di Milano
}

\date{}

\maketitle

\begin{abstract}
Learning Classifier Systems (LCS) are population-based reinforcement learners used in a wide variety of applications.  This paper presents a LCS where each traditional rule is represented by a spiking neural network, a type of network with dynamic internal state.  We employ a constructivist model of growth of both neurons and dendrites that realise flexible learning by evolving structures of sufficient complexity to solve a well-known problem involving continuous, real-valued inputs.  Additionally, we extend the system to enable temporal state decomposition. By allowing our LCS to chain together sequences of heterogeneous actions into macro-actions, it is shown to perform optimally in a problem where traditional methods can fail to find a solution in  a reasonable amount of time.  Our final system is tested on a simulated robotics platform.

\end{abstract}

\section{Introduction}

Many real-world processes encompass some temporal element that describes or affects the state of a given system through time.  Artificial Intelligence ubiquitously utilises abstractions of real-world or natural systems, however many representations opt to remove these temporal facets during the abstraction process.  Missing out on underlying temporal patterns in a problem potentially denies a system access to a rich vein of information which can be harnessed to generate more efficient solutions.

Within machine learning, temporal problems are particularly prevalent in the field of robot control, whereby robotic agents act through a number of time-steps to achieve some user-defined goal.  To achieve any kind of reasonably complex behaviour, temporal aspects of the environment must be taken into account.  This is perhaps most easily realised by using some temporally-dependent solution representation to match the nature of the problem representation.  In this work, we concentrate on Spiking Neural Networks (SNNs), which are temporally-sensitive abstractions of biological neural networks. They are of particular interest to us as (i) they have been successfully applied to myriad robotic control problems (ii) there is growing evidence that the more biologically-realistic spiking models are efficient forms of knowledge representation (e.g. \cite{maass}, \cite{sag-wex}).  

Our specific approach to temporal machine learning involves the use of a Learning Classifier System, or LCS \citep{Holland76}, a form of online evolutionary reinforcement-based learning that evolves a population of {\em(condition, action, prediction)} rules using a Genetic Algorithm (GA) \citep{Holland75}. 

Traditionally, LCSs have used a ternary rule condition structure, comprising binary digits \{0,1\} and a generalisation character \{\#\} to match in certain subspaces of a binary-represented state space.  However, such a representation limits the application of LCSs to problems which can be binary-represented.  Rule representations have since been extended to handle integer  and real-valued states using several representations including intervals (\cite{Wilson1999b} and \cite{Wilson2001b} respectively), hyperellipsoids \citep{lanzi:2006:hyp} and convex hulls \citep{lanzi:2006:ch}, opening LCSs to more varied problem domains.  

Further extensions enlarged the remit of LCS, including LISP S-expression rule conditions \citep{lanzi:1999:ERCCPIFMCS}, computed actions \citep{Ahluwalia1999}, and fuzzy logic with computed actions  \citep{Valenzuela-Rendon1991a}. Artificial neural networks have also been used to the same effect \citep{Bull:2002:UCN}.  We follow previous work with LCS using neural classifiers, (e.g. \cite{conf/gecco/HowardBL08}), which has shown that it is possible to allow evolution to control both the number of hidden layer nodes (termed Neural Constructivism) and the number of node connections (termed Connection Selection) during the reinforcement learning process, alongside each classifier's rates of mutation. In particular, the work has utilized multi-layered perceptrons (MLP) \citep{rumelhart86} to generate optimal action selection policies by calculating actions based on the input state.

To date, only one scheme that can utilise temporal information has been used within LCS; a form of recurrent Boolean logic network \citep{Bull2009a}. Here, we combine a modern LCS with a temporally dynamic SNN representation by replacing each classifier condition/action pair with a SNN. It is shown that the use of a SNN allows the classifiers to harness the network's persistent, dynamic internal state to produce temporally-dependant activation patterns.  Additionally, we modify our LCS as in \citep{conf/cec/StudleyB05} to create a Temporal Classifier System (TCS), which facilitates generalisation in both time and space, thereby enabling more direct use of the temporal behaviour of the neurons for determining actions.

Our hypothesis is that  (i) the evolved SNN networks will display significant benefits compared to previous work using MLP network \citep{howard-gecco09} (ii) the use of temporally-sensitive SNN classifiers will extend the remit of XCSF to solving a class of problems that include an element of time e.g. semi-Markov Decision Processes (MDPs) \citep{sutton-99}.

Specifically, we address the following research questions: does the use of SNN networks compared to MLP networks provide any significant advantage in terms of performance, solution size or network parsimony?  Does the evolutionary design process evolve the networks in different ways?  How well does the system compare to a benchmark reinforcement learner?  Is there synergy between TCS and SNN classifiers, in that TCS provides a framework that allows the SNNs to harness their temporal element more effectively, which in turn increases the performance of TCS?

To address these questions, we test our SNN representation against both an MLP representation and a tabular Q-learner \citep{watkins-thesis}, which acts as a benchmark, on a standard Reinforcement Learning (RL) \citep{SaB} test problem (a 2-D grid world).  We then modify the environment to include a temporal element and test our SNN-TCS against an MLP-TCS, to demonstrate the utility of the former network representation over the latter.  One final modification increases the difficulty of the environment by increasing the amount of steps it takes to reach the goal state; again we compare to a Q-learner to demonstrate the ability of TCS to generate optimal action policies in a more difficult environment by using underlying temporal information in the problem.  It is further shown that this temporal, spiking LCS performs optimally in a more challenging, realistic simulated robotics task.

The main contributions of this paper are (i) the combination of spiking networks and a temporally-sensitive, self-adaptive and constructive LCS framework, which are shown to perform optimally in continuous time and space where tabular Q-learning performs poorly (ii) the subsequent deployment of the aforementioned system in a simulated robotics task, demonstrating the power of the system in a nontrivial semi-MDP.



\section{Neural Representations}

The initial work exploring artificial neural networks within LCS used traditional feedforward MLPs to represent the rules \citep{Bull:2002:UCN}. Recurrent MLPs were then shown able to provide memory for a simple maze task \citep{bull-hurst-tech03}. Radial Basis Function (RBF) networks (e.g. \cite{Buhmann}) were later used for both real \citep{journals/alife/HurstB06} and simulated robotics \citep{BullO02} tasks.  Analysis of the rule sets produced  in \citep{BullO02} shows networks evolve that compute different actions that vary depending upon the environmental input, as long as the correct payoff value for those $(state, action)$ combinations is identical. The authors hypothesize that it may be possible for the system to function optimally with only a single neural rule per possible payoff value, attesting to the compact rule representation and generalization capabilities of the neural LCS.  Both MLP and RBF networks have been shown amenable to a constructionist approach, which can be defined as a method whereby the number of nodes within the hidden layer is under evolutionary control, along with the connection weights (\cite{Bull:2002:UCN}; \cite{journals/alife/HurstB06}).  We have recently extended this vein of research by introducing explicit network-wide feature selection, termed  ``connection selection'', which allows each inter-neural connection to be probabilistically enabled or disabled, with the aim of reducing the number of connection weights within the rules \citep{conf/gecco/HowardB08}.  

The use of spiking networks is based on the assumption that simpler network forms can constrain the types of solutions evolved by the system in a manner that is detrimental to performance or solution parsimony, especially when considering temporal factors.  Two well-known formal spiking implementations are the Leaky Integrate and Fire (LIF) model and the Spike Response Model (SRM).  They are covered in detail in \citep{spiking-n-m}.  A number of previous studies have evolved single spiking networks.  Neuroevolution (the use of evolution to design neural achitectures) was first applied to spiking networks by \cite{Korkin:1998:SII}, in this case evolving networks that produce temporally-dependant outputs.  SRM networks were later evolved for vision-based robot navigation \citep{Floreano:2001:ESN}.  The authors conclude that the inherent dynamics of a spiking network may be suited to certain robotics tasks.  A survey of various methods for evolving both connection weights and architectures in neural networks is presented in \citep{flor-durr-matt}.  To our knowledge, the work presented in this article is the first approach to evolving spiking neurons that includes both variable network sizes, connection selection and self-adaptive parameters within an “ensemble” of networks (population of classifiers).  By combining these elements within XCSF, we are able to demonstrate generalization performance in excess of our previous MLP-based systems.

\section{XCSF}

XCSF comprises a population of classifiers, the main components of which are the {\em condition},  {\em action} and {\em prediction}.  The {\em condition} is used to determine whether or not the classifier matches the current $s_t$, the {\em action} is the action the classifier advocates, and the {\em prediction} is the predicted payoff the classifier expects from carrying out its action given the state $s_t$.  Each classifier also keeps track of myriad other parameters, including the last time it was involved in GA activity, the accuracy of its prediction value, and the fitness of the classifier.  For further details, we refer the interested reader to \citep{Wilson2001a}.

Within XCSF,  the fitness of a classifier is related to the accuracy of its prediction of payoffs. This leads to a system where all areas (not just high reward areas) of the problem landscape are covered by classifiers that predict expected payoff with a high degree of accuracy. XCSF further increases its generalization capabilities by computing predicted payoffs. That is, classifier prediction is not a constant value as in other LCS, such as XCS \citep{wilson:1995:cfba}. Rather, prediction is calculated linearly as the product of the sensory input and a prediction weight vector, which allows the same classifier to generalize by predicting different payoff values in different areas of the environment.   At the start of each experiment, each classifier is initialized with a prediction weight vector $w$, used to compute the classifiers prediction. This vector has one element for each input, plus an additional element $w_0$ which corresponds to $j_0$, a constant input that is set as a parameter of  XCSF.  Each prediction weight vector element is initially 0. 

XCSF involves two varieties of trial, exploration and exploitation, which are carried out with approximately equal frequency.  At each time-step, XCSF builds a match set, [M], from [P] consisting of all classifiers whose conditions match the current input state $s_t$.  In the binary case, each classifier condition has one element per state variable, where state variables are binary and classifier conditions are a ternary string  augmented with  \#, where \# is a generalisation character that matches any state input at that position.  A classifier is said to match if it has either (i) the same binary digit as the corresponding state  or (ii)\# at each position in its condition.  In traditional XCSF, each action must be present in each [M].   If this is not the case, {\em covering} is used to generate classifiers that advocate the missing action(s); {\em covering} generates a new classifier whose condition is a copy of the  input state, with generalisation characters probabilistically inserted.  Once [M] is formed, the prediction array is created.  Classifier prediction ($cl.p$) is calculated linearly as a product of the environmental input (or state, $s_t$) and the prediction weight vector ($w$) associated with each classifier, shown in equation 1.  

\begin {equation}
 cl.p(s_t) = j_0 + \sum_{i>0} cl.w_i * s_{t(i)} \label{pred1}
\end {equation}

The prediction array is the fitness-weighted average of the calculated predictions for each possible action. The prediction array is then used to decide on an action to take (in traditional XCSF this is deterministic during an exploit trial and random during an explore trial).  Once an action is selected, all classifiers that advocate the selected action form the action set [A]. The action is taken and, if the goal state is reached, a reward is returned from the environment that is used to update the parameters of the classifiers in [A].  A discounted reward is propagated to the previous action set [A$_{-1}$] if it exists. During reinforcement, rather than updating the classifier's prediction value, the prediction weight vector of each classifier in the action set is updated using a version of the delta rule.  Prediction error is then updated as in \citep{Wilson2001a}.




XCSF makes use of macroclassifiers to enhance computational efficiency.  A macroclassifier represents a number of virtual single classifiers, or microclassifiers, and its numerosity indicates the number of identical microclassifiers that the macroclassifier represents. This provides a computational advantage over representing each microclassifier individually.  Subsumption is the mechanism by which macroclassifiers are formed; XCSF uses two forms of subsumption.  GA subsumption checks whether a child condition is already represented by more general parent, and increases the parents' numerosity by one if this is the case.  Action set subsumption checks the most general classifier in [A] against all other members of [A], and increases its numerosity by one if its condition is more general than theirs.  In either scenario, the subsumed classifier is deleted from [P].  The GA may then activate if the average time since the last GA application to the classifiers in [A] exceeds a threshold $\theta_{GA}$.  Traditionally in XCSF, two offspring classifiers are generated by reproducing, crossing, and mutating the parents. The offspring are inserted into the population, two classifiers are deleted if the maximum population is reached. As happens in all other models of classifier systems, parents stay in the population competing with their offspring.

\section {Neural XCSF}

Transitioning to a neural classifier representation (N-XCSF) requires modifications to be made to the algorithm detailed in Sect. 3.  It should be noted that the use of neural classifiers only alters the matching, {\em covering} and action selection mechanisms.  It is also important to note the distinction between ``connection weight'', which refers to a node-to-node weight within a neural network and ``prediction weight'', which is used as in XCSF to calculate classifier prediction linearly (after \cite{Wilson2001a}).

By using a neural network to replace the condition, and calculate the matching and action of a classifier, we increase generalisation by allowing the classifier to advocate heterogeneous actions in different regions of the problem subspace.  The input state, $s_t$, is used by the classifier in two regards: (i) to calculate whether the classifier matches in the problem subspace, and determine the action it advocates if it does match (ii) to calculate prediction values for the classifier in a given problem subspace in which it matches.

\begin{figure}[h!]
\begin{center}
\centerline{ 
\psfig{ file=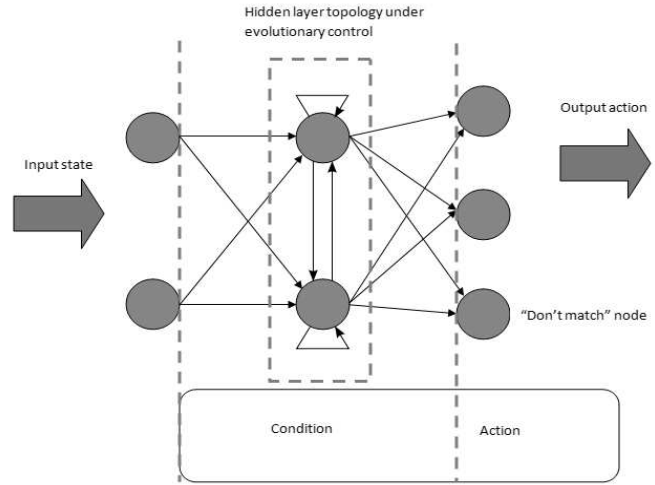,width=9cm, height=7cm}
}
\end{center}
\caption{Detailing the mapping between a neural classifier and a traditional classifier.  A legal connectivity pattern for a SNN classifier is shown.}
\label{fig:spike-classifier}
\end{figure}

For all experiments considered here, output neurons (real-valued numbers in the case of MLPs, spike trains in the case of SNNs) must be discretised into having  either {\em high} or {\em low} activation for the purposes of action calculation.  Each network comprises a problem-dependent number of input neurons, a number of hidden layer neurons under evolutionary control (see Sect. 5.1), and three output neurons. The first two output neurons are used to calculate the action.  Similarly to \cite{Bull:2002:UCN}, the final output neuron is a “don’t-match” neuron that excludes the classifier from the match set if it is highly activated.  This is necessary as the action of the classifier must be re-calculated for each state the classifier encounters, i.e. a classifier may advocate different actions in different regions of state space.  Covering is altered to repeatedly generate random networks until the network action matches the desired output for a given input state.   Fig. ~\ref{fig:spike-classifier} shows how a neural classifier relates to a traditional classifier condition and action.

\subsection{MLP Classifiers}

The MLP network is a highly abstracted neural model that can produce real-valued outputs from real-valued inputs.  Each neuron uses a sum-over-inputs, constrained by a sigmoid function, to produce an output value.  Each classifier condition/action pair is represented by a vector that is realized as a feed-forward MLP.  Each weight in this condition is uniformly initialized randomly in the range [-1, 1] so that a node can have both positive and negative postsynaptic connections.  Networks are initially fully-connected.  MLP networks cannot have recurrent connections, nor can they have connections within the same layer.  High activation is  $>$  0.5, low is $\leq$0.5. 

\subsection{SNN Classifiers}

Spiking neural networks present a biologically plausible phenomenological model of information processing in the brain.  Mathematical models of the most basic current spiking neuron, the Leaky Integrate and Fire (LIF) neuron, can be traced back to as early as 1907 \citep{lapique}.  We base our spiking implementation on the LIF model, although it must be stressed that our model is heavily simplified in terms of the number of simulation steps used per action calculation.  Newly initialized networks are fully connected, and both recurrent connections and hidden-to-hidden layer connections are legal.  Each connection weight is uniformly initialized randomly in the range [0, 1] and cannot become negative via GA action, as the type of node determines the nature (positive or negative) of the connection.  Hidden layer nodes are initially excitory with 50\% probability, otherwise they are inhibitory.  All input and output layer nodes are excitory.  Action calculation involves the current input state being processed five times by each network.  Each output neuron has an activation window that records the spike train produced by that neuron over these five steps.  To calculate actions, we classify an output spike train as {\em highly activated} if it has spikes in over half of the positions in the sampling window after five runs (3 - 5 spikes); otherwise it is said to have low activation.   

Neurons can be stimulated either by an external current or by spikes from other connected neurons.  Each neuron has an internal state (or membrane potential, $m$) which acts as a measure of its current activation level.  As spikes are received by the neuron, the value of $m$ is altered.   If $m$ surpasses a threshold, $mthresh$,  the neuron sends a spike with value 1 to all connected neurons, who then receive a reduced current relative to the connection weight between those two neurons.  A “leak” constant is used to gradually reduce $m$  to 0 over time ($m$ is always $>$0).  At time $t$, the membrane potential of a neuron is given in equation 2:

\begin {equation}
 m(t+1) = m(t) + (I + a - bm(t))
\end {equation}
\begin {equation}
if (m(t) > mthresh)  m(t) = c
\end {equation}

Equation 3 shows the reset formula.  Here, $m(t)$  is the membrane potential at time $t$, $I$ is the input current to the neuron, $a$ is a positive constant, $b$ is the degradation (leak) constant and $c$ is the reset membrane potential of the neuron. Normally, spiking neurons have their initial membrane potentials $m$ set to the membrane reset value $c$.  We employ a bootstrapping mechanism whereby the initial membrane potential of every network is set to $c\_ini$ rather than $c$, where $c\_ini>c$.  This allows the networks to achieve their first spike quicker,  giving less “setup” cycles as the networks begin to respond  more strongly to the inputs, allowing us to (i) use smaller window sizes at the output neurons - a performance-enhancing measure (ii) benefit from faster general emergence of temporally-sensitive activation patterns within the networks.  Note that after an initial spike, neurons are reset to $c$ and not $c\_ini$; boot-strapping only affects the time to first spike of each neuron in the network.

\subsection{Discovery Component}

In N-XCSF, the GA is modified to be a two-stage process.  Stage 1, detailed in the following paragraph,  controls the rates of mutation and constructivism/connection selection that occur within the system, with stage 2 (Sect. 5) controlling the evolution of neural architecture in terms of both neurons and connections.  In N-XCSF we use mutation exclusively to explore connection weight space; crossover is omitted as experimental evidence thus far suggests that, in neural classifier systems \citep{conf/gecco/HowardBL08}, sufficient solution space exploration can be obtained via a combination of self-adaptive connection weight and topology mutations (see also \citep{journals/ijon/RochaCN07}). However, we note that a number of suitable neural crossover operators have previously been presented (e.g. by \cite{NEAT}).  A GA is periodically triggered in [A] to evolve fitter classifiers in an environmental niche. We utilise self-adaptive mutation rates as in \citep{Bull2000d}, to dynamically control the amount of genetic search, or frequency of mutation events, taking place within the niche. This provides stability to areas of the problem space that are already ``solved'' as the mutation rate for a niche is typically directly proportional to its distance from the goal state during learning; generalization learning, along with the value function learning, occurs faster nearer the goal state. Here, the $\mu$ value (rate of mutation per allele) of each classifier is initialized uniformly randomly in the range [0,1]. During a GA cycle, a parent`s $\mu$ value is modified as in equation 4 - the offspring then adopts this new $\mu$, with $(0< \mu \leq1)$, mutates itself by this value, and is inserted into the population:

\begin {equation}
\mu \leftarrow \mu * e^{N(0,1)}
\end {equation}

\section{Topology Mechanisms for Neural Networks}
A classic problem in neural networks revolves around network topology considerations; how many neurons should a network consist of?  How should we configure their topological arrangement and inter-neural connectivity patterns to ensure acceptable performance?  In addition to self-adaptive mutation, in this work two evolutionary topology morphology schemes are applied to allow the modification of the spiking networks in two regards: by adding/removing hidden layer neurons; and adding/removing inter-neural connections.  This allows each classifier to control its own knowledge representation autonomously in terms of both frequency and range of mutation that takes place in a given niche at a given time \citep{Bull2000d}, and by adapting the hidden layer topology of the neural networks to reflect the complexity of the problem sub-space considered by the network \citep{Bull:2002:UCN}.  

\subsection{Constructivism}
Constructivist learning (e.g., \citep{QandS}) postulates that neural structures are initially small and sparsely connected.  Learning acts to add appropriate structure, in the form of neurons and connections, until some satisfactory level of computing power is attained; suitable specialized neural structures emerge as a result of the learner’s interaction with its environment.  The implementation of constructivism in this system is based on the aforementioned work in neural LCS \citep{Bull:2002:UCN}.  Each rule has a varying number of hidden layer neurons (initially 1, and always $>$ 0), with additional neurons being added or removed from the single hidden layer depending on the constructivism element of the system. 

Constructivism takes place during a GA cycle, after mutation. Two new self-adaptive parameters, $\psi$ $(0<\psi \leq1)$ and $\omega$ $(0<\omega \leq1)$, are added. Here, $\psi$ represents the probability of performing a constructivism event and $\omega$ is the probability of adding a neuron, with removal occurring with probability ($1-\omega$). As with self-adaptive mutation, both are initially randomly generated uniformly in the range [0,1], and offspring classifiers have their parents $\psi$ and $\omega$ values modified during reproduction as with $\mu$.  SNN nodes created during constructivism are initially excitatory with 50\% probability sampled from a uniform distribution, otherwise they are inhibitory.  We have previously shown the utility of this approach in contrast to using fixed-size networks (e.g. \cite{conf/gecco/HowardBL08}).

\subsection{Connection Selection}
Feature selection (FS) is a method of reducing the number of the data inputs to a process by selecting and operating exclusively on a subset of those inputs.  A popular FS variant in the machine learning community is automatic FS. Automatic FS includes both wrapper approaches (where feature subsets can be tailored during the running of the algorithm, e.g., \cite{KohaviJohn:97}) and filter approaches (where subset selection is a pre-processing step and subsets are immutable during the running of the algorithm, e.g. \cite{conf/icml/KollerS96}).  The use of FS in a neural context corresponds to enabling/disabling connections between the input and hidden layer, and brings two major benefits.  Firstly, the amount of data being input to a process can be reduced (increasing computational efficiency), secondly noisy connections (or those otherwise inhibitory to the successful performance of the system) can be disabled.  The connection structure of artificial neural networks was first evolved by Dolan and Dyer \cite{Dolan1987}.  A comparative summary can be found in \cite{conf/ecal/SchlessingerBL05}, where many neuro-evolution methodologies are compared by the authors. Explicit FS within LCS has been investigated by \cite{bull-adam-07} and \cite{BacarditSHSLK07}. \

In this paper we allow any connection to be individually enabled/disabled, a mechanism termed ``Connection Selection''.  Connection selection is implemented in our system as follows: During a GA cycle, and based on a new self-adaptive parameter $\tau$ $(0<\tau \leq1)$ (which is initialized and self-adapted in the same manner as the other parameters), an enabled connection can be disabled, or vice versa.  If a connection is enabled, its connection weight is randomly initialised uniformly in the range [0, 1].  All connections are initially enabled for new classifiers and classifiers created via cover.  During a node addition event, new connections are enabled probabilistically, with {\em P(connection enabled)} = 0.5, with connection weight randomly set uniformly in the range [0,1] as before.  We previously have shown the utility of this approach for reducing network complexity (e.g., \cite{conf/gecco/HowardB08}).

For clarity, we now summarise the steps involved in a GA cycle.  First, two offspring networks are selected.  The self-adaptive parameters for those networks are altered as in equation 4.  Connection weights are altered based on $\mu$, then node addition/removal takes place based on $\psi$ and $\omega$.  Finally, connections are added or removed based on $\tau$.  These networks are inserted into the population and networks are deleted as required. 

\section{Experiments in Continuous State Space}

Experiments were conducted on the continuous 2-D grid world \citep{boyan.moore-1995:gener}, a standard continuous Reinforcement Learning (RL) \citep{SaB} test problem. We compare to our earlier work using an MLP-based  N-XCSF (see \cite{conf/gecco/HowardBL08}; \cite{howard-gecco09} for details).  Each experiment was comprised of a number of trials.  A trial was defined as starting when the agent is initially positioned randomly in the environment, consisting of a number of [M] and [A] formations as the agent navigated in the environment, and finishing either with the agent reaching the goal state and receiving reward, or the trial timing out after the agent has moved 200 steps (a time-saving measure).  Each trial was either in exploration mode (roulette wheel action selection, see e.g. \cite{conf/gecco/HowardBL08} for reasoning) or exploitation mode (deterministic action selection). 

\subsection{Continuous Grid World}

In the Grid World, the agent's current state, $s_t$, is defined as its $(x,y)$ position, each bounded in the range [0,1]; any movement outside of this range takes the agent to the nearest grid boundary.  Increasing environmental complexity, and to emulate the sensory noise encountered in the real world, both the perceived $x$ and $y$ position of the agent are subject to noise; +/- [0\%-5\%] of the actual position.  The agent moves a predetermined step size (0.05) in one of four directions (North, East, South, West).  The goal state is in the top-right hand corner of the grid – where $(x+y >1.90)$.  The agent is intiallty placed anywhere except the goal, and must navigate to the goal in the fewest possible steps (the average optimum is 18.6 steps), whereupon it receives an immediate reward of 1000 and the next trial begins. All other actions give an immediate reward of 0.  The environmental discount rate, $\gamma$, is set to 0.95.

Each experiment consisted of 20,000 trials, 10,000 explore and 10,000 exploit.  Experiments were repeated ten times with the results being the mean average of these 10 runs.  At the end of each exploitation trial we performed an additional exploitation trial from an arbitrary fixed location far from the goal state (0.25, 0.25) to determine when the system reached a stable solution.  This location is also subject to noise; +/- [0\%-5\%] of the actual agent position.  We defined the system to be stable if, for 50 consecutive trials, it provided an optimal steps-to-goal value when starting from the selected location.  That measure of stability allowed us to perform standard t-tests to compare the respective performances of the two neural representations.

The current state of the system was analyzed every 50 trials and used to create the figures.  In the ``steps-to-goal'' figures, the dashed line indicates optimal performance, all averages are averages from [P].    In the following tables, ``Stability'', is the average time the system takes to reach stability.  ``Connectivity'' is defined as the average percentage of enabled connections per classifier in [P].  ``Mutation'' is the average final self-adaptive $\mu$ parameter in [P]. ``Neurons'' are the average final connected hidden layer neurons per classifier in [P].   ``Macroclassifiers'' shows the final number of macroclassifiers contained in [P]. 

Action calculation proceeds as follows: the outputs at the two output neurons (not the ''don't match`` node)  are mapped to a single discrete movement in one of four compass directions (North, East, South, West). Four possible directions and only two ranges of discrete output are possible: low and high.  The combined actions of the neurons translate to a discrete movement according to the two motor output strengths – (high, high) = North, (high, low) = East, (low, high) = South, and (low, low) = West.

\begin{figure*}[t!]
\begin{center}

\subfloat[]{ \psfig{file=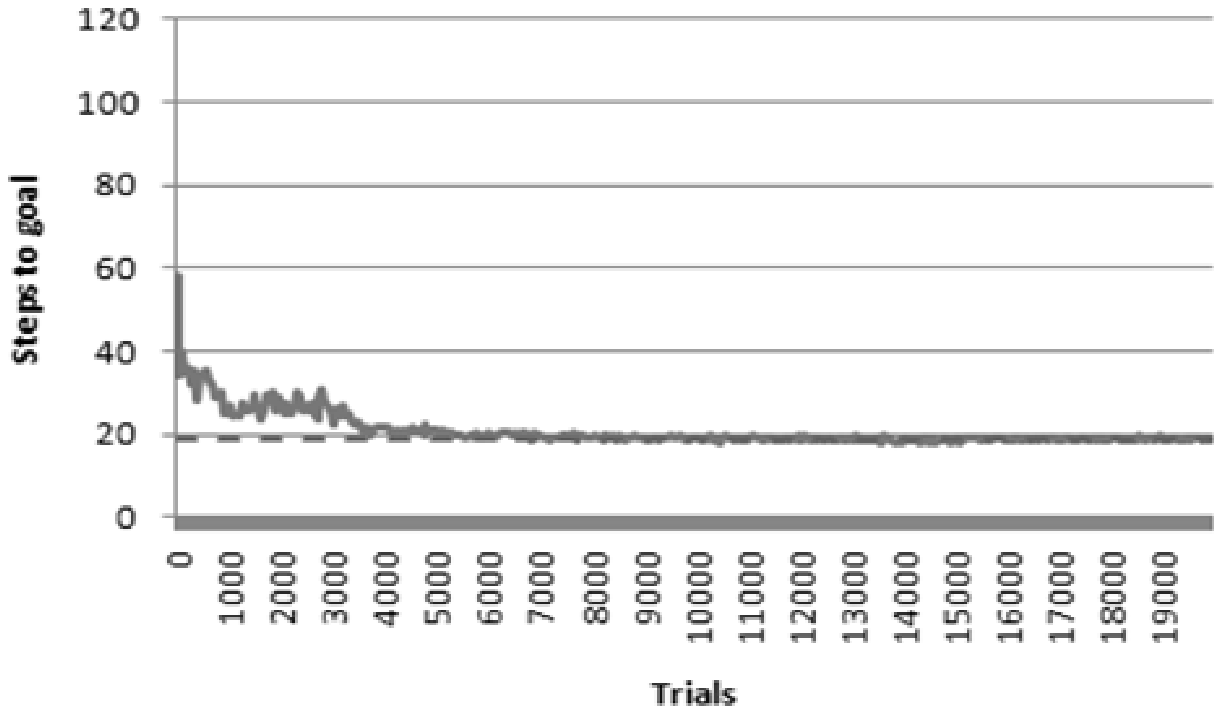,width=8cm,height=4cm}}
\subfloat[]{ \psfig{file=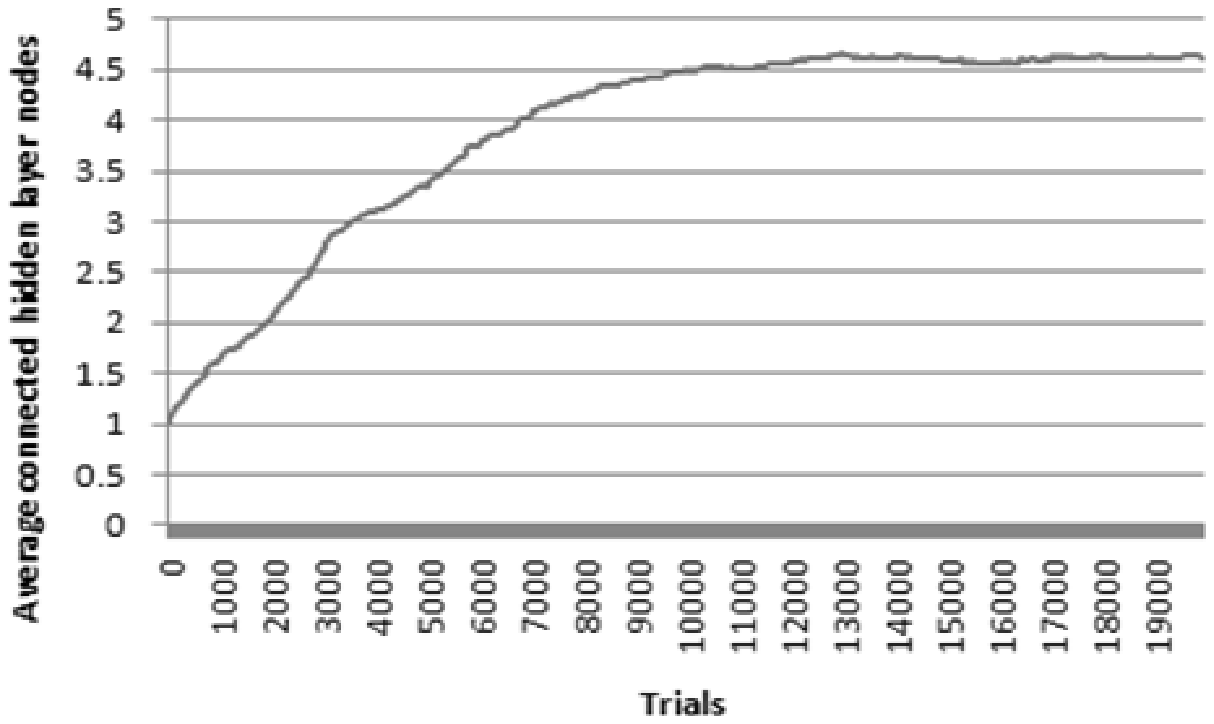,width=8cm,height=4cm}}\\
\subfloat[]{ \psfig{file=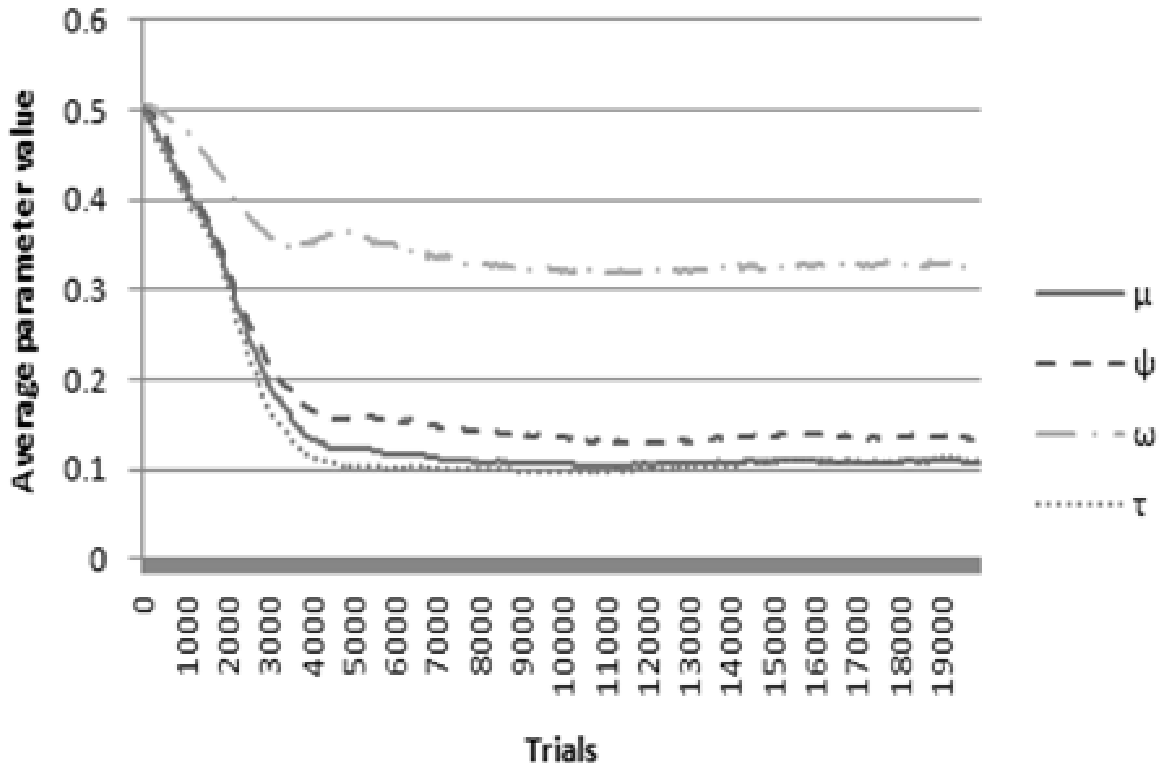,width=8cm,height=4cm}}
\subfloat[]{ \psfig{file=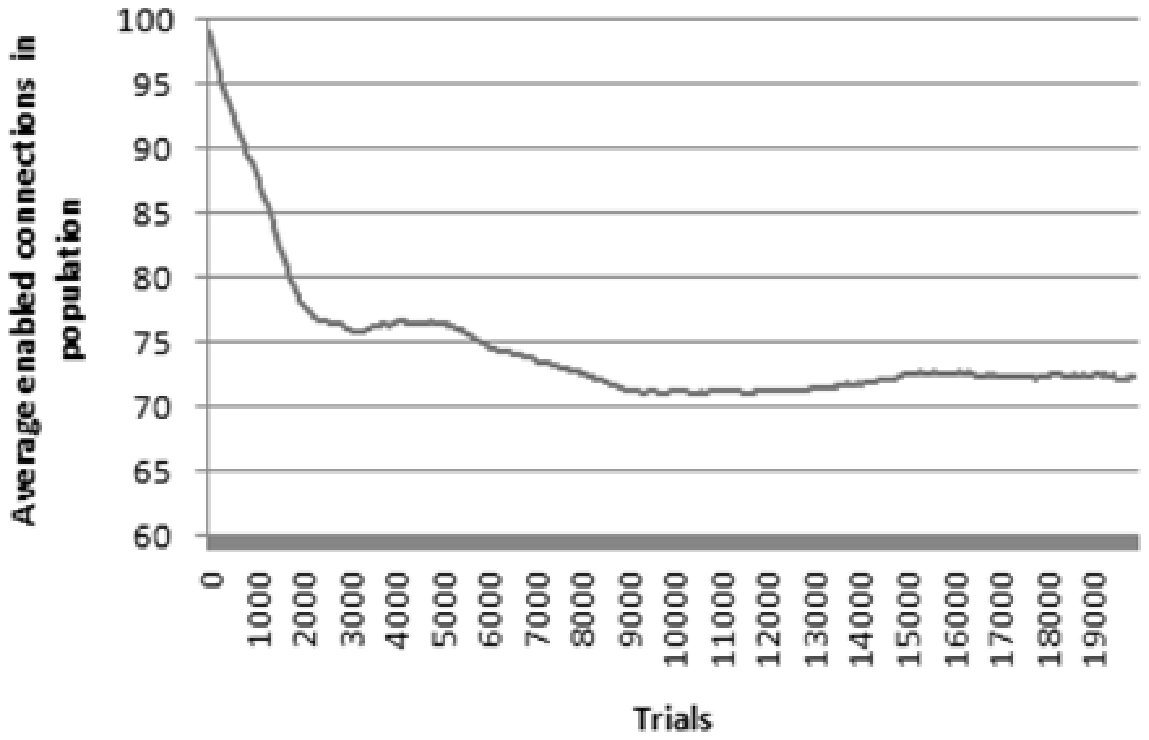,width=8cm,height=4cm}}

\end{center}
\caption[]{Continuous grid world (a) Steps to goal, (b) average connected hidden layer nodes, (c) average self-adaptive parameter values, (d) average enabled connections in spiking N-XCSF}
\label{Spike}
\end{figure*}

\begin{figure*}[t!]
\begin{center}

\subfloat[]{ \psfig{file=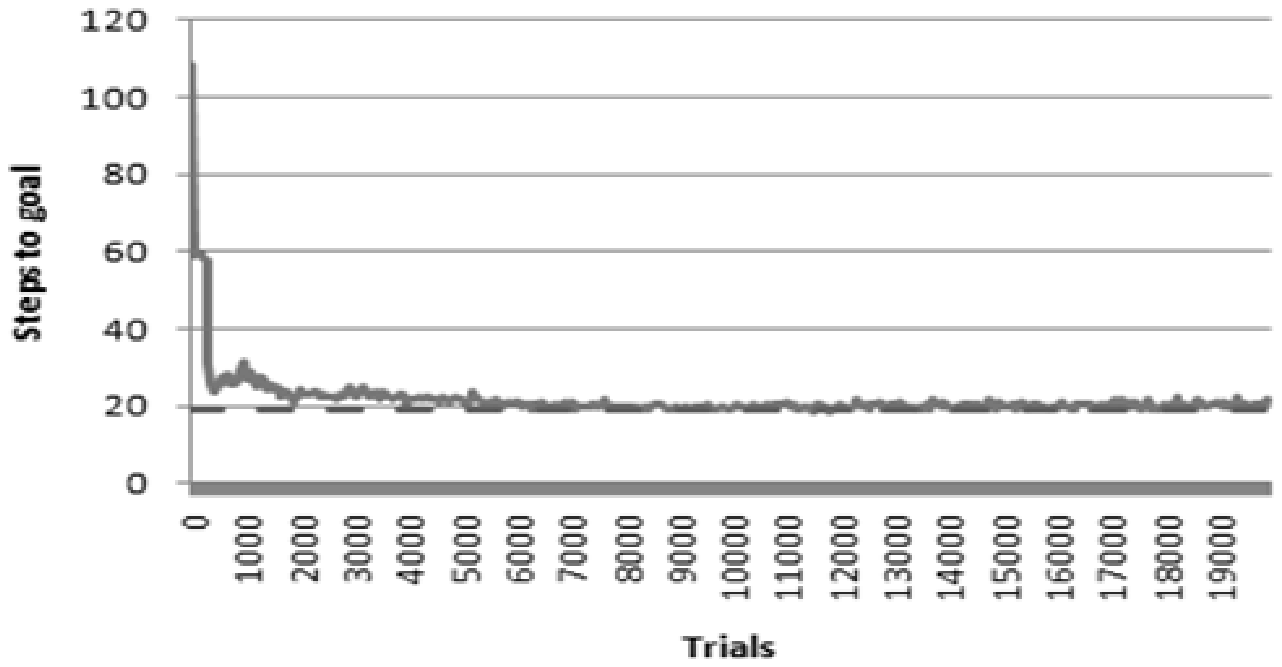,width=8cm,height=4cm}}
\subfloat[]{ \psfig{file=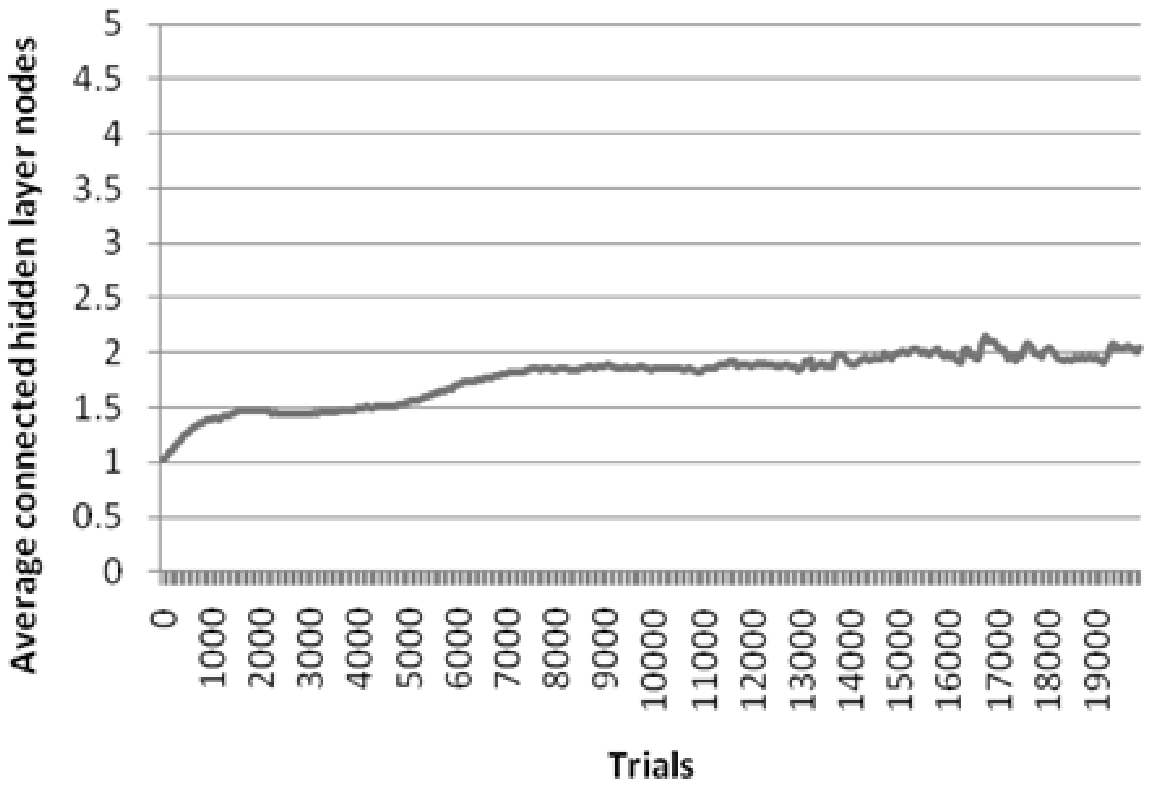,width=8cm,height=4cm}}\\
\subfloat[]{ \psfig{file=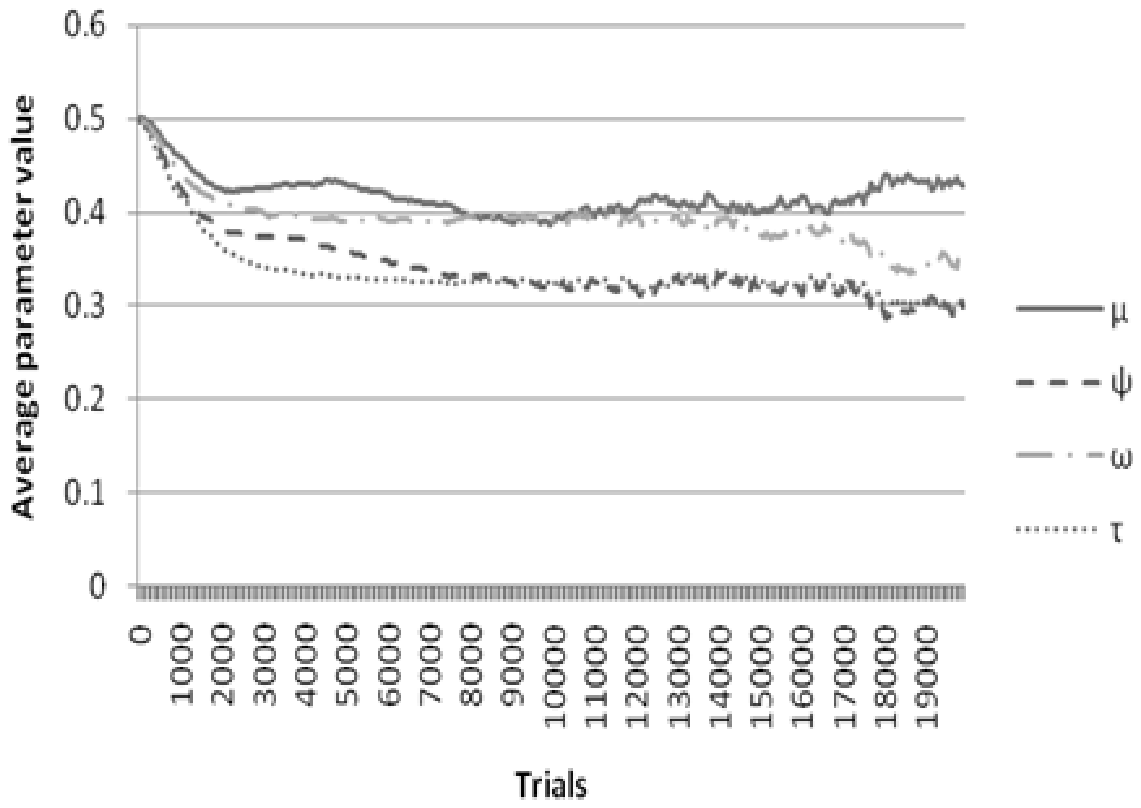,width=8cm,height=4cm}}
\subfloat[]{ \psfig{file=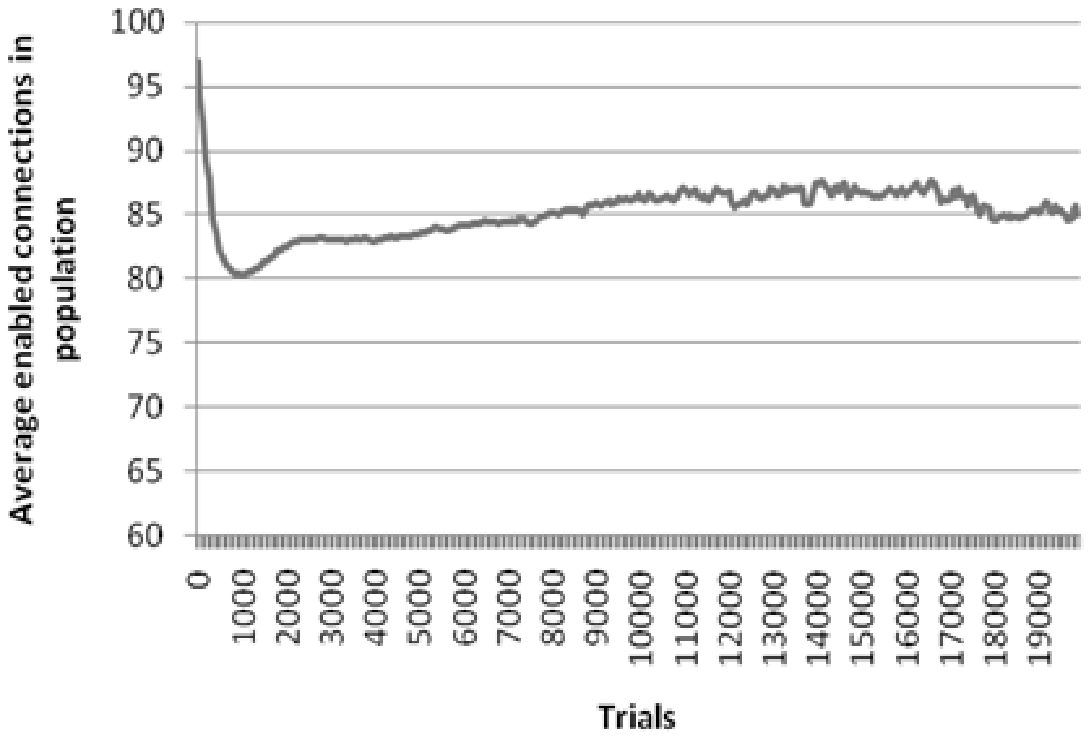,width=8cm,height=4cm}}

\end{center}
\caption[]{Continuous grid world (a) Steps to goal, (b) average connected hidden layer nodes, (c) average self-adaptive parameter values, (d) average enabled connections in MLP N-XCSF}
\label{MLP}
\end{figure*}

Spiking N-XCSF was parameterized as follows: population size $N$=20000, learning rate $\beta$=0.2 $ (0<\beta \leq1)$, accuracy threshold $\epsilon_0$=0.005,  GA threshold $\theta _{GA}$=50, deletion threshold $\theta _{DEL}$=50, XCSF constant $x_0$=1, XCSF learning rate $\eta$=0.2 $(0<\eta \leq1)$.  All other XCSF parameters follow \citep{Wilson2001a}.  Spiking parameters are $a$=0.3, $b$=0.05, $c$=0, $c\_ini$=0.5, $mthresh$=1.0 and {\em output window size}=5.  All networks initially have a single hidden layer neuron.

Table ~\ref{tab:MLP-spike} shows t-test results when comparing the two network representations in solving the continuous grid world.  The first row shows that time to stability is not statistically significantly affected by network type (p-value 0.13) (although the spiking version does have a lower mean average – also compare Figs. ~\ref{MLP}(a) and ~\ref{Spike}(a)).  The internal problem representation, in terms of both self-adaptive mutation rate (p=8.6$\times$10$^{-10}$) and number of neurons added via constructivism , (p=1.5$\times$10$^{-4}$) is formed in a statistically significantly different manner.  As is typical in our constructivist approach, we note context sensitive structures being formed during the learning process, in two main ways (i) where the context is the network type: certain topological arrangements being favoured by either SNN or MLP networks and appearing more frequently in the final solutions (ii) where the context is the spatial position: the evolution of similar ``processing groups'' of neurons at certain locations in the problem space, within solutions of the same network type.  Perhaps the most striking result is that the number of macroclassifiers used is significantly lower in the SNN case (p=1.08$\times$10$^{-6}$), indicating that spiking networks are capable generating more compact overall solutions.

Although more nodes are required for a spiking representation (Fig.~\ref{MLP}(b) shows 2.01 connected nodes, Fig.~\ref{Spike}(b) shows 4.62 connected nodes), the spiking networks are less connected (p=5$\times$10$^{-4}$;  Fig.~\ref{MLP}(d) shows 85\% connectivity, Fig.~\ref{Spike}(d) shows 72\%).  Self-adaptive parameter values are lower in the spiking case (Fig. ~\ref{MLP}(c) values range between 0.3 and 0.43,  Fig.~\ref{Spike}(c) values range between 0.1 and 0.32), especially the three parameters concerned with actual rates of mutation (e.g. $\mu$, $\psi$, $\tau$).  These universally lower mutation rates indicate that a more stable solution is being evolved by the spiking networks, in terms of the classifiers used in the final solution having less chance of mutation, and less allelle variation per mutation event, in the SNN case.  The more uniform curvature observed in Figs.~\ref{Spike}(b) and~\ref{Spike}(c), as opposed to Figs.~\ref{MLP}(b) and~\ref{MLP}(c), indicates that not only is the final solution more stable, the evolution process itself is made more stable.  Spiking performance seems to compare favourably to an interval representation (e.g. \citep{conf/cec/LanziLWG05a}, without noise), especially since the spiking networks produce compact solutions in the presence of sensory noise.

\begin{table}
\begin{center}
\caption[]{Detailing t-test results in MLP and spiking versions of N-XCSF in the continuous grid world}
\label{tab:MLP-spike}
\begin{tabular}{|l|l|l|l|} 
\hline Metric & Network type & Average & P-value \\ 
\hline Stability & MLP & 11453.5 &  \\ 
  & Spiking & 8280.5 & 0.13 \\ 
\hline Mutation & MLP & 0.45 &  \\ 
 & Spiking & 0.108 & 8.6$\times$10$^{-10}$ \\ 
\hline Neurons & MLP& 2.01 & \\
 & Spiking& 4.623&1.5$\times$10$^{-4}$ \\
\hline Connectivity &MLP& 84.95& \\
&Spiking &72.33 & 5$\times$10$^{-4}$\\
\hline Macroclassifiers&MLP&11237.3 &  \\
 &Spiking &3006.9 &1.08$\times$10$^{-6}$  \\
\hline 
\end{tabular}
\end{center}
\end{table}

We additionally compared spiking N-XCSF to a tabular Q-learner (as in \citep{conf/gecco/LanziLWG05a} for XCSF) with discount rate $\gamma$ = 0.99.  We performed the optimal spatial discretisation (in terms of Q-learner performance) as given in \citep{loiacono-thesis} to discretise the continuous space into a 21 $\times$ 21 grid for the $x$ and $y$ coordinates that comprise $s_t$, as shown in equation 5.  The purpose of this experiment, as well as the experiment presented towards the end of Sect. 8,  is to provide an indicator of the performance of our system against a baseline reinforcement learner.

\begin {equation}
disc\_x = (int)(floor (cont\_x \times 1/(step\_size)))
\end {equation}

Q-learning rapidly achieves optimal performance within 500 trials.  This result significantly outperforms our spiking N-XCSF (p=3.2$\times$10$^{-3}$, with an average time to stability of 8280.5 in the spiking N-XCSF case, compared to 82.6 for the tabular Q-learner).  However, Q-learning requires a suitable discretisation of state space to be decided on beforehand. The Q-value learned is identical to the payoff value learned in XCSF, the difference being the use of function approximation in the latter.

\section{Taking Time into Consideration}

Reinforcement learning methods typically assign a value to each possible state-action combination of a given task. When a programmer is prepared or able to define the state space discretisation a priori, this methodology has been proven to work for robot systems \citep{Mahadevan/Connell:1992}. The approach is however labour intensive and becomes less tractable when the state space complexity increases. There are numerous accepted methods by which generalization of state spaces can be achieved, two of the most popular being gradient descent methods and linear approximation (specifically tile coding) (e.g., \citep{conf/icml/Lin91}, \citep{journals/ras/Tham95}). \cite{santa} extended the latter approach by considering continuous-duration action spaces.  Within LCS, the approach has traditionally been either to predefine the temporal duration of an action \citep{journals/ai/DorigoC94} or to predefine the amount by which a sensor reading must change before a change in state is said to have occurred \citep{Cliff1994a}.

Motivation for adding actions of a continuous duration is to more closely bridge the gap between simulation and physical implementation;  an agent samples its sensors with low latency, and reacts accordingly.  Our goal is to go some way to model these rapid sensory updates by facilitating them in our learning architecture.  In this case, we employ a system whereby an action set can control the agent for more than one {\em continuous-duration action}, which can be comprised of many {\em discrete actions}, updating the computed action for the classifiers in [A] as new environmental states present themselves.  The LCS can therefore create high-level continuous macro actions from chains of lower-level actions.  As the LCS can search the space of possible macro actions, the need to predefine such actions is removed. 

Our system is based on the “Temporal” LCS (TCS) which has been used with both ZCS (originally \citep{Wilson94}, implemented in \citep{Hurst:2002:TLC}) and, more recently, with XCS \citep{conf/cec/StudleyB05} and with XCSF and MLPs \citep{howard-gecco09}. In TCS, the match set [M] and the action set [A] are formed as usual.  Subsequent input states are then fed into [A] as the agent traverses the environment, without reforming [M] as in normal LCS.  If all classifiers in [A] still match the input, the advocated action is taken and next input state retrieved. If no classifiers in the current [A] match the newly presented input state, or a timeout limit since the [M] formation is reached, the action is dropped (with reinforcement and GA activity) and a new match set formed as normal.  If only some classifiers match, [A] is split into two new sets, [C] (the continue set) and [D] (the drop set), so that each classifier has dual-set membership; [A] and either [C] (if it still matches) or [D] (if it does not).  Roulette wheel selection based on fitness-weighted predicted payoff is then used to pick a classifier from [A], and its membership of either [C] or [D] is used to determine the whether the system continues or drops the current action.

\begin {itemize} 
\item If [C] wins we continue, removing [D] classifiers from [A].
\item If [D] wins we remove [C] classifiers from [A], perform necessary parameter updates, then drop [A] and form a new [M].  
\end {itemize}

As we continually calculate actions in [A], some classifiers may advocate different actions to the one originally used to comprise [A] from [M].  In this case, a ``winning'' action is picked from [A] based on the action selection policy of the trial (roulette during exploration, deterministic during exploitation).  All classifiers that do not advocate the newly-selected action are removed from [A] but not added to the drop set; as the outcome of using their advocated action is not explored, an accurate prediction value cannot be ascertained.  Traditionally, reinforcement in LCS is given with the formula in equation 6.  Here, $r$ is the immediate reward, $\gamma$ is the discount factor and $maxP$ is the maximum of the prediction array. The reinforcement update is altered to consider the amount of time an [A] maintains control of the LCS and the global time taken to achieve reward:

\begin {equation}
P=r+ \gamma \times maxP
\end {equation}

\begin {equation}
P= (e^{- \varphi t^t}) r + (e^{-\rho t^i}) \times maxP
\end {equation}

Equation 7 shows the TCS reinforcement forumula; the first and second reward factors, $e^{- \varphi t^t}$  and $e^{- \rho t^i}$, favour efficient overall solutions (in terms of the overall number of discrete actions) and efficient state transitions (between continuous-duration actions) respectively, $\varphi$  and $\rho$ are experimentally-determined discounting factors, $t^t$ is the total number of steps taken in the trial, and $t^i$ is the number of steps taken since the last match set formation.  

\section{Experiments in Continuous Time and Space}

In the following experiments we prove the ability of SNN to handle continous time as well as continuous space.  We validate our approach on two well-known RL test problems, the Grid World described in Sect. 7 and the ``mountain-car'' problem.  It is important to note that we only reset the network membrane potentials ($m$) for every node once at the beginning of each trial.  In this way, we hoped to exploit the temporal dynamics inherent to spiking neural representations to solve these continuous time, continuous space environments more effectively by preserving network states between action set formations.  In contrast to experiments in Sect. 7, the consideration of time classes the problems as semi-MDPs.

\subsection{Continuous Grid World}

\begin{figure*}[t!]
\begin{center}

\subfloat[]{ \psfig{file=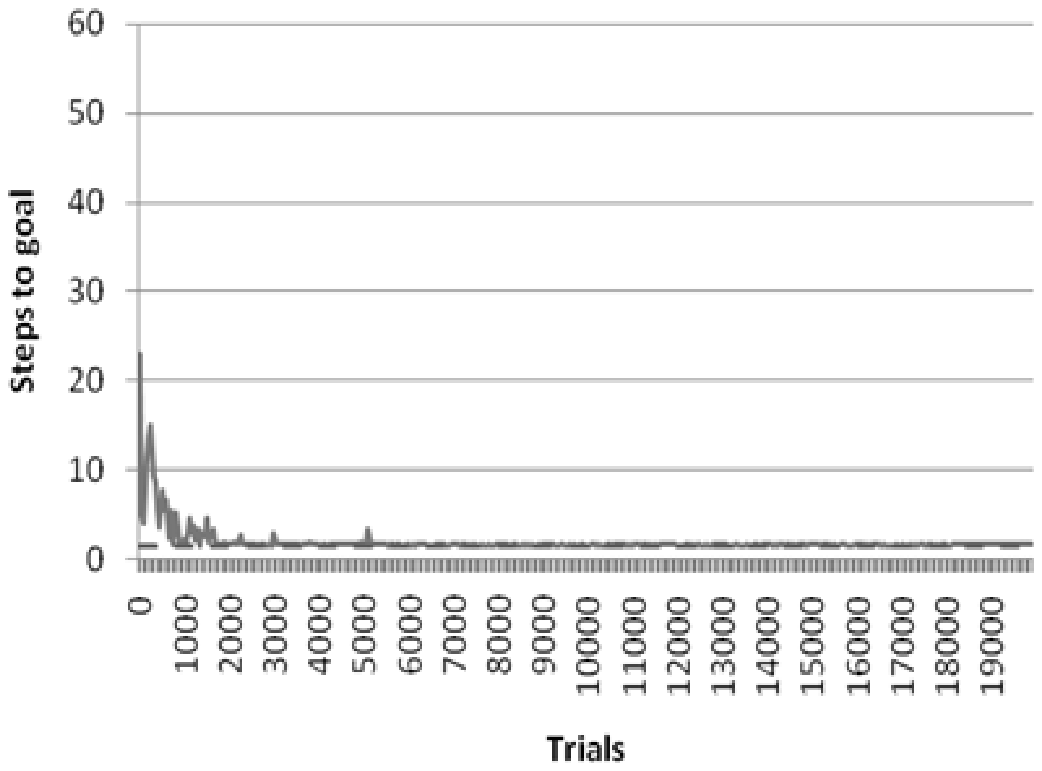,width=8cm,height=4cm}}
\subfloat[]{ \psfig{file=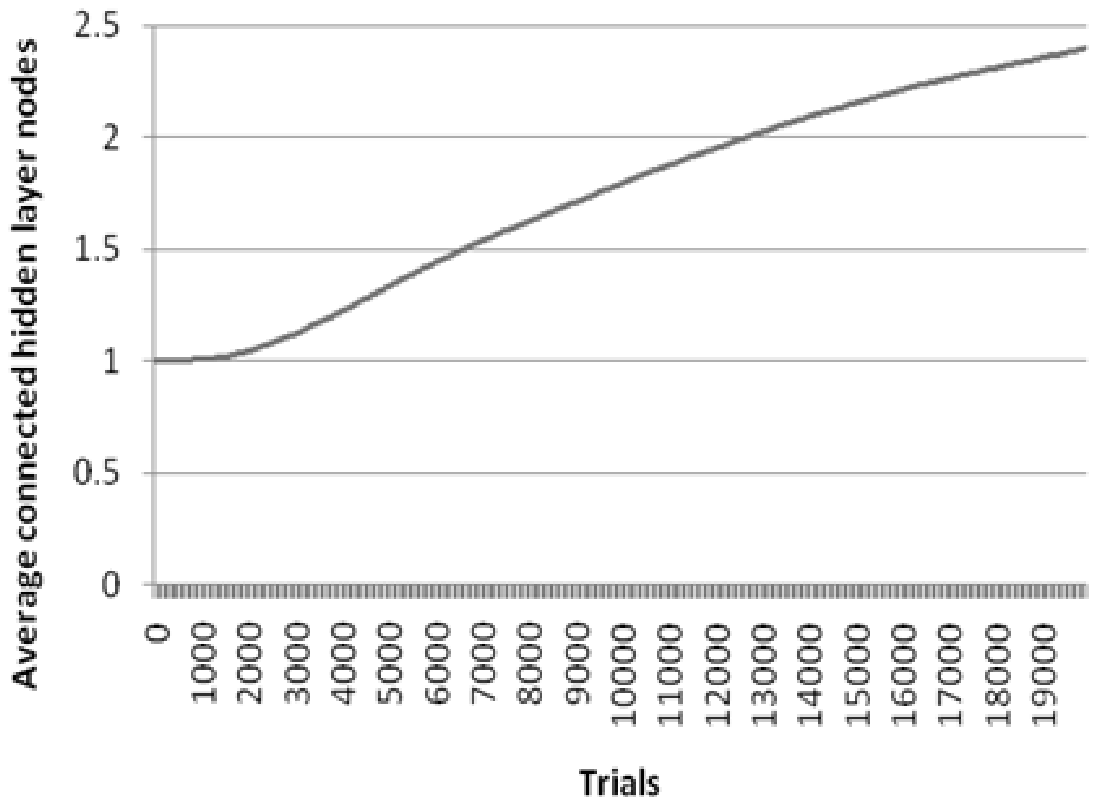,width=8cm,height=4cm}}\\
\subfloat[]{ \psfig{file=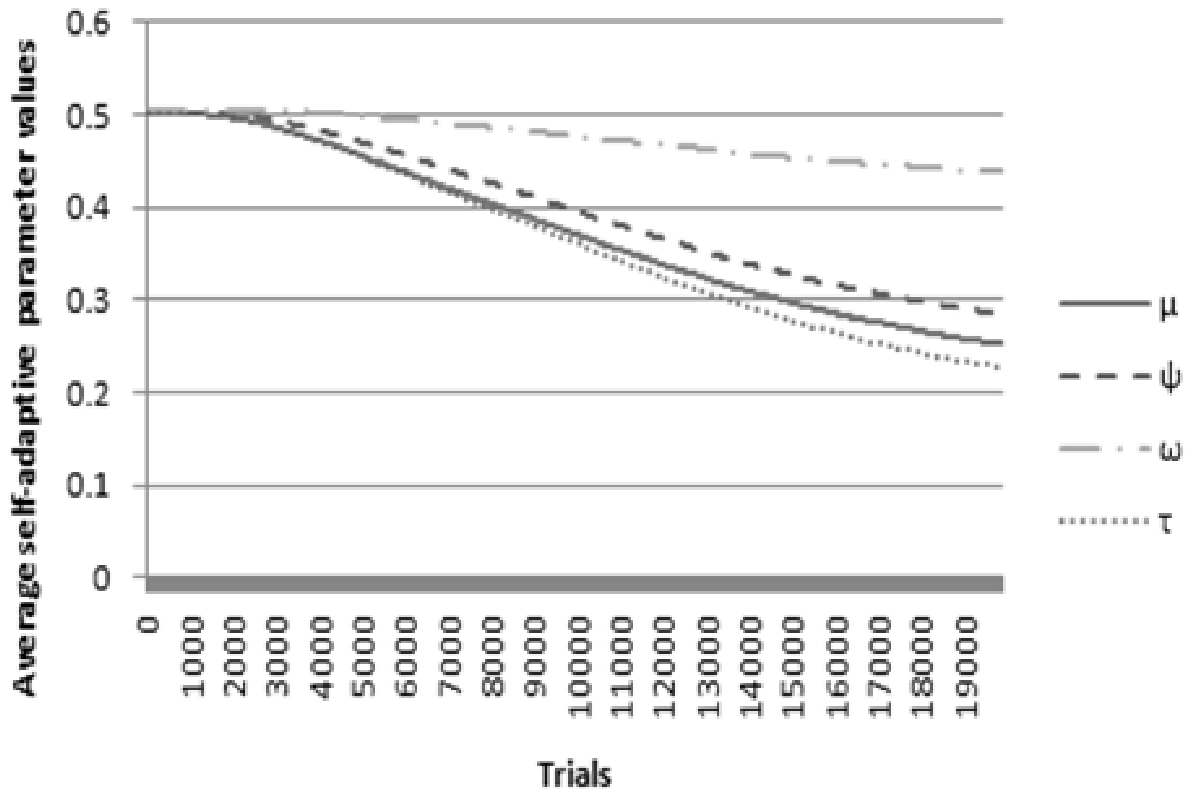,width=8cm,height=4cm}}
\subfloat[]{ \psfig{file=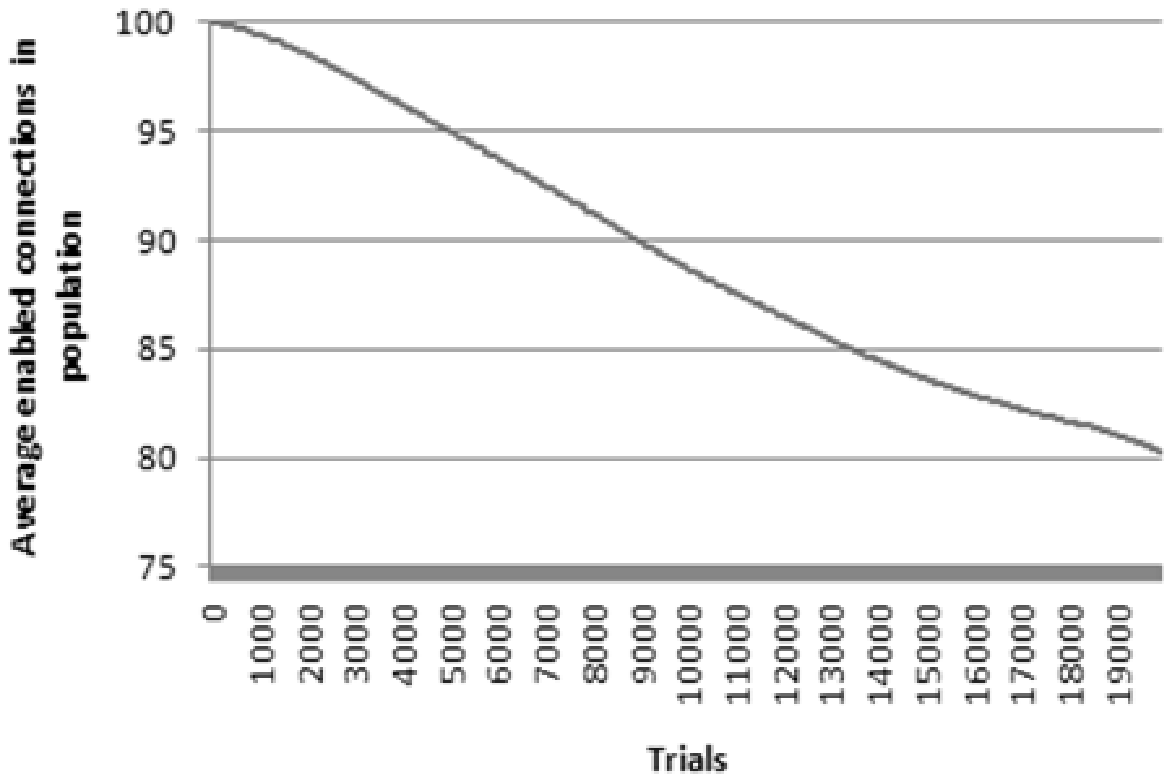,width=8cm,height=4cm}}

\end{center}
\caption[]{Continuous grid world (a) Steps to goal, (b) average connected hidden layer nodes, (c) average self-adaptive parameter values, (d) average enabled connections in spiking TCS N-XCSF}
\label{Spike-TCS}
\end{figure*}

\begin{figure*}[t!]
\begin{center}

\subfloat[]{ \psfig{file=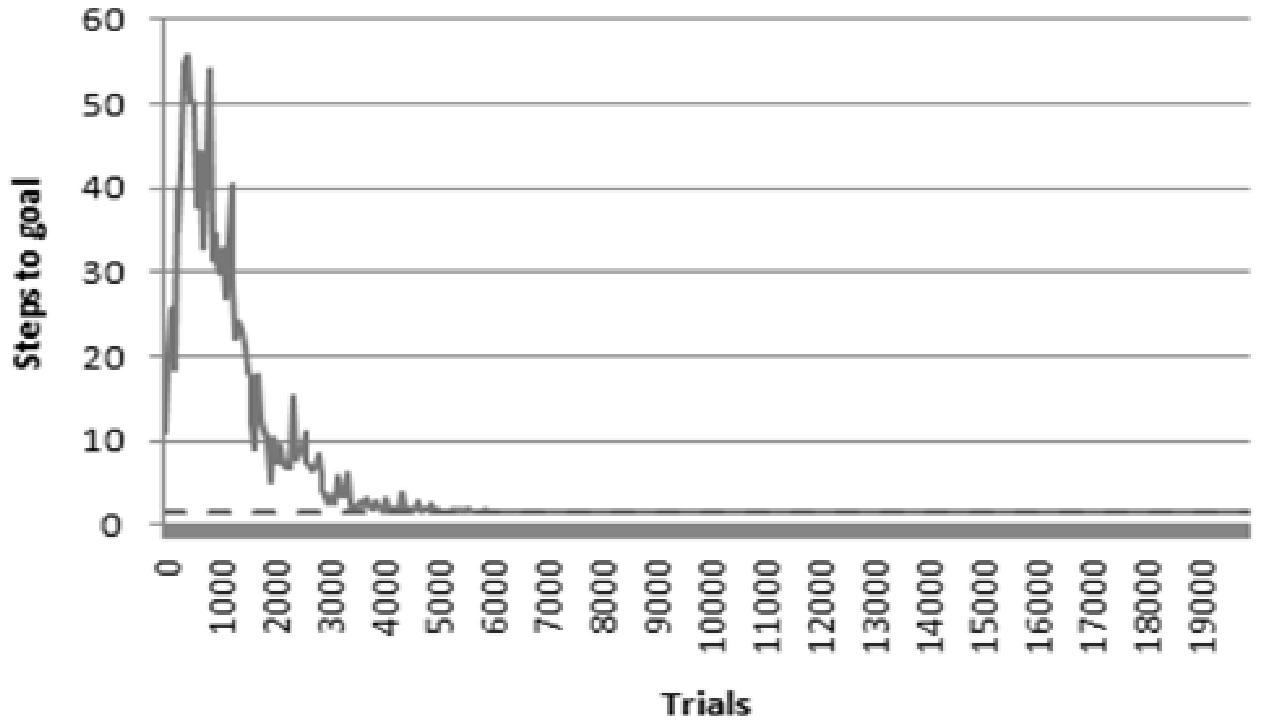,width=8cm,height=4cm}}
\subfloat[]{ \psfig{file=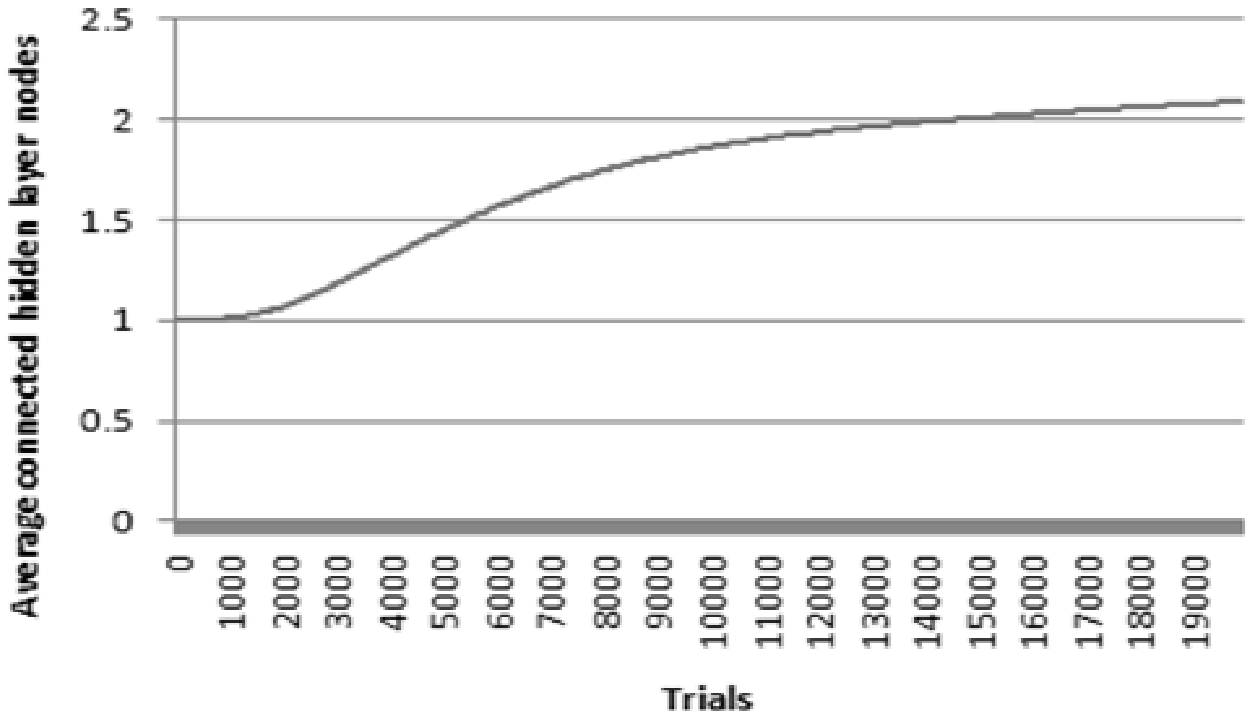,width=8cm,height=4cm}}\\
\subfloat[]{ \psfig{file=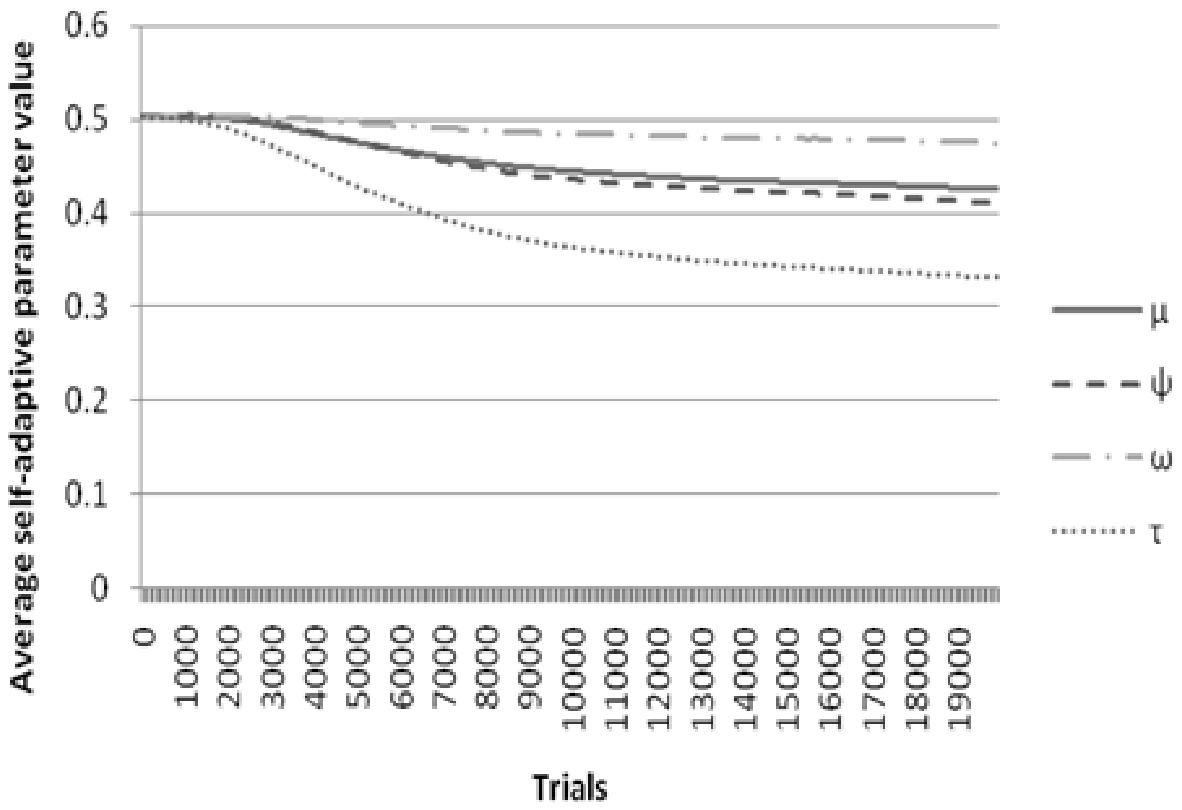,width=8cm,height=4cm}}
\subfloat[]{ \psfig{file=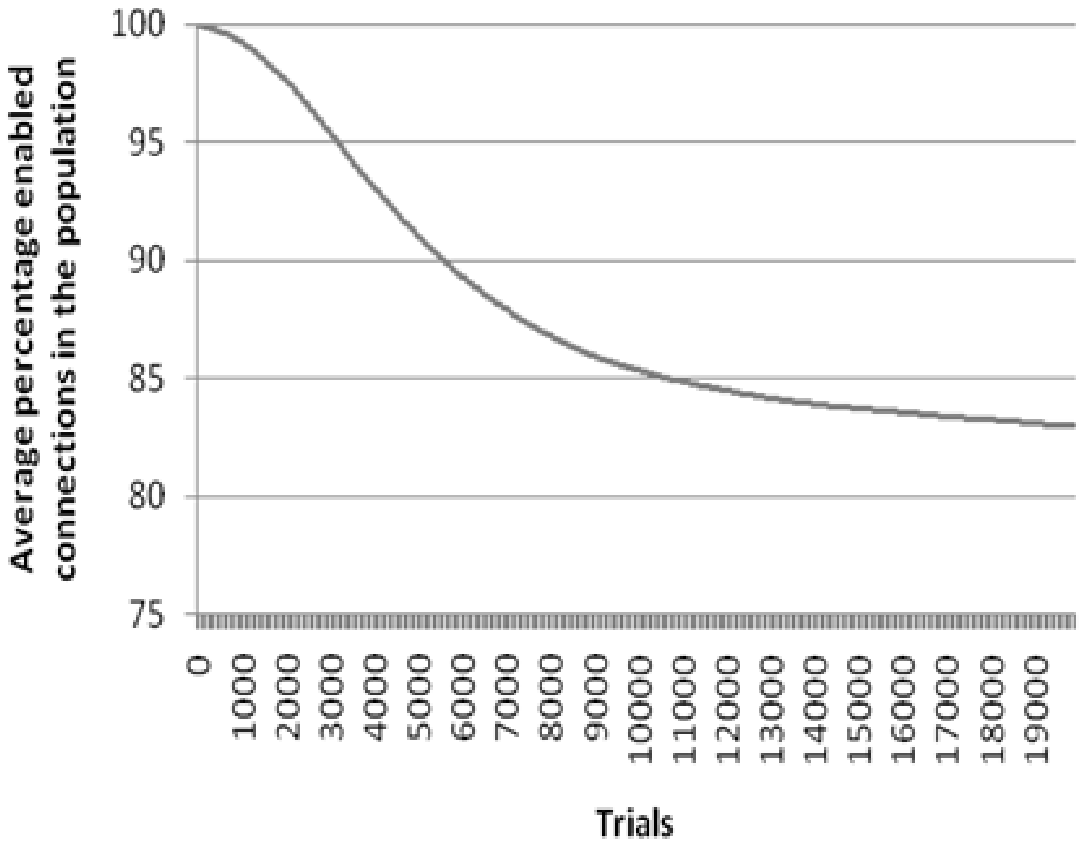,width=8cm,height=4cm}}

\end{center}
\caption[]{Continuous grid world (a) Steps to goal, (b) average connected hidden layer nodes, (c) average self-adaptive parameter values, (d) average enabled connections in MLP TCS N-XCSF}
\label{MLP-TCS}
\end{figure*}

In the Grid World,  TCS parameters were set as:  $\varphi$=0.45, $\rho$=0.005, $timeout$=20.  All other parameters were identical to those in Sect. 7.  As with the previous experiments involving the Grid World, we compare our SNN TCS implementation to an MLP-based TCS.  Table~\ref{MLP-spike-TCS} describes the t-test results for SNN and MLP implementations; showing no significant performance difference between the two neural representations (p=0.57).  Fig.~\ref{Spike-TCS}(a) shows optimal performance after 2000 trials for SNN, compared to 5000 trials in the MLP case, which is shown in Fig.~\ref{MLP-TCS}(a). No significant differences are found in terms of connected hidden layer nodes (p=0.51). Fig.~\ref{Spike-TCS}(b) shows 2.4 connected hidden layer nodes, and Fig.~\ref{MLP-TCS}(b) shows just over 2 connected hidden layer nodes, indicating that the SNN representation requires larger hidden layer sizes on average. In terms of both mutation stability (p=1.5$\times$10$^{-4}$) and number of macroclassifiers (p=1.5$\times$10$^{-3}$) the spiking version held the advantage in a statistically significant manner, attesting to the benefits of a SNN representation over an MLP one.  In terms of self-adaptive parameters, Fig.~\ref{Spike-TCS}(c) values ranged between 0.22 and 0.44 and Fig.~\ref{MLP-TCS}(c) values ranged from 0.48 to 0.33 - all self-adaptive parameters were lower in the spiking case; again, an indicator of improved stability in terms of less mutation events on average in the evolved SNN classifiers. 

Figs.~\ref{Spike-TCS}(d) and~\ref{MLP-TCS}(d) show the differences in connectivity between SNN (86\%) and MLP (84\%) networks; revealing the spiking to be more densely connected, although not statistically significantly so (p=0.065).  It is also interesting to note that although the transition to TCS in the spiking case did not statistically increase solution stability it decreased the average number of neurons required by the networks with a p-value of 0.016, highlighting the ability of a SNN to produce temporally dynamic activity, as fewer processing units were required when the temporally-sensitive SNNs were coupled with a semi-MDP test problem.  This indicated that the SNNs were harnessing temporal information contained in the test problem to solve it in a different manner.

\begin{figure}[t!]
\begin{center}

 \psfig{file=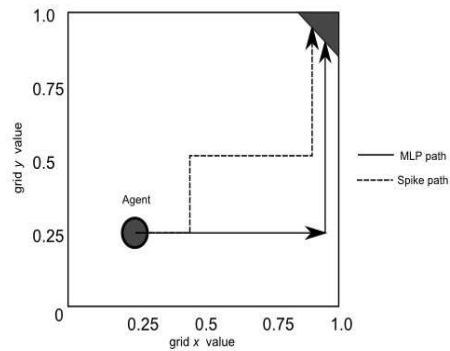,width=6.0cm,height=5cm}

\end{center}
\caption[]{Sample action discretisations in spiking and MLP networks in the continuous grid world}
\label{spike-vs-mlp-moves}
\end{figure}

It was interesting to observe that the spiking networks seemed more predisposed to allow for multiple actions to be selected within a single match set formation, as opposed to the more homogenous action selection evidenced in MLP networks (e.g. \citep{conf/gecco/HowardBL08}).  Fig.~\ref{spike-vs-mlp-moves} shows heterogeneous action selection in the spiking case allowing a single match set to control the agent all the way to the goal state, contrasting with the MLP solution of more homogenous action selection; first taking the agent to the rightmost border, then progressing up the border to the goal state.  In situations where both network types provided heterogeneous actions from the same match set, the spiking networks were more likely to sequentially switch selected action from one step to the next.  Fig.~\ref{spike-vs-mlp-moves} shows that, given the actions {North, South, East, West} in the continuous grid world, the spiking representation gave (N, E, N, E, N, E...) whereas the MLP representation provided (N, N, N... E, E, E...).  The more complex discretisations available to the spiking representation gives a clear benefit of using a spiking representation over an MLP representation, when considering the prospective application of the system to more complex environments where highly varied, heterogeneous action selection may be required for optimal performance.

\begin{table}
\begin{center}
\caption[]{Detailing t-test results in TCS-enabled MLP and spiking versions of N-XCSF in the continuous grid world}
\label{MLP-spike-TCS}
\begin{tabular}{|l|l|l|l|} 
\hline Metric & Network type & Average & P-value \\ 
\hline Stability & MLP & 5687.5 &  \\ 
  & Spiking & 5637.67 &0.57 \\ 
\hline Mutation & MLP & 0.43 &  \\ 
 & Spiking & 0.25 & 1.5$\times$10$^{-4}$ \\ 
\hline Neurons & MLP& 2.09 & \\
 & Spiking& 2.40&0.51 \\
\hline Connectivity &MLP& 83.93 & \\
&Spiking &86.13 & 0.065 \\
\hline Macroclassifiers&MLP&18148.1 &  \\
 &Spiking &16552.25 &1.5$\times$10$^{-3}$  \\
\hline 
\end{tabular}
\end{center}
\end{table}

\subsection {Smaller Step Size Environment}

In preparation for a move to a simulated robotics platform, we decreased the step size tenfold to 0.005.  Motivation for this experimentation was seen as testing the scalability of our temporal representation, as a real robot would read sensors many times per second. The new step size was more akin to the number of readings a real robot may be required to make.  It may also be necessary for the agent to perform homogenous action selection in extended regions of the environment, followed by certain highly heterogeneous areas where more complex behavioural policies, such as obstacle avoidance, are required; TCS allows for discretisation of state space based on required actions, as well as potentially altering the action of a given [A] through multiple state transitions in response to the required action transition frequency.  Finally, a smaller step size introduced an environment where a TCS classifier system should be able to perform optimally, whereas a traditional Q-learner struggled.

\begin{figure*}[t!]
\begin{center}

\subfloat[]{ \psfig{file=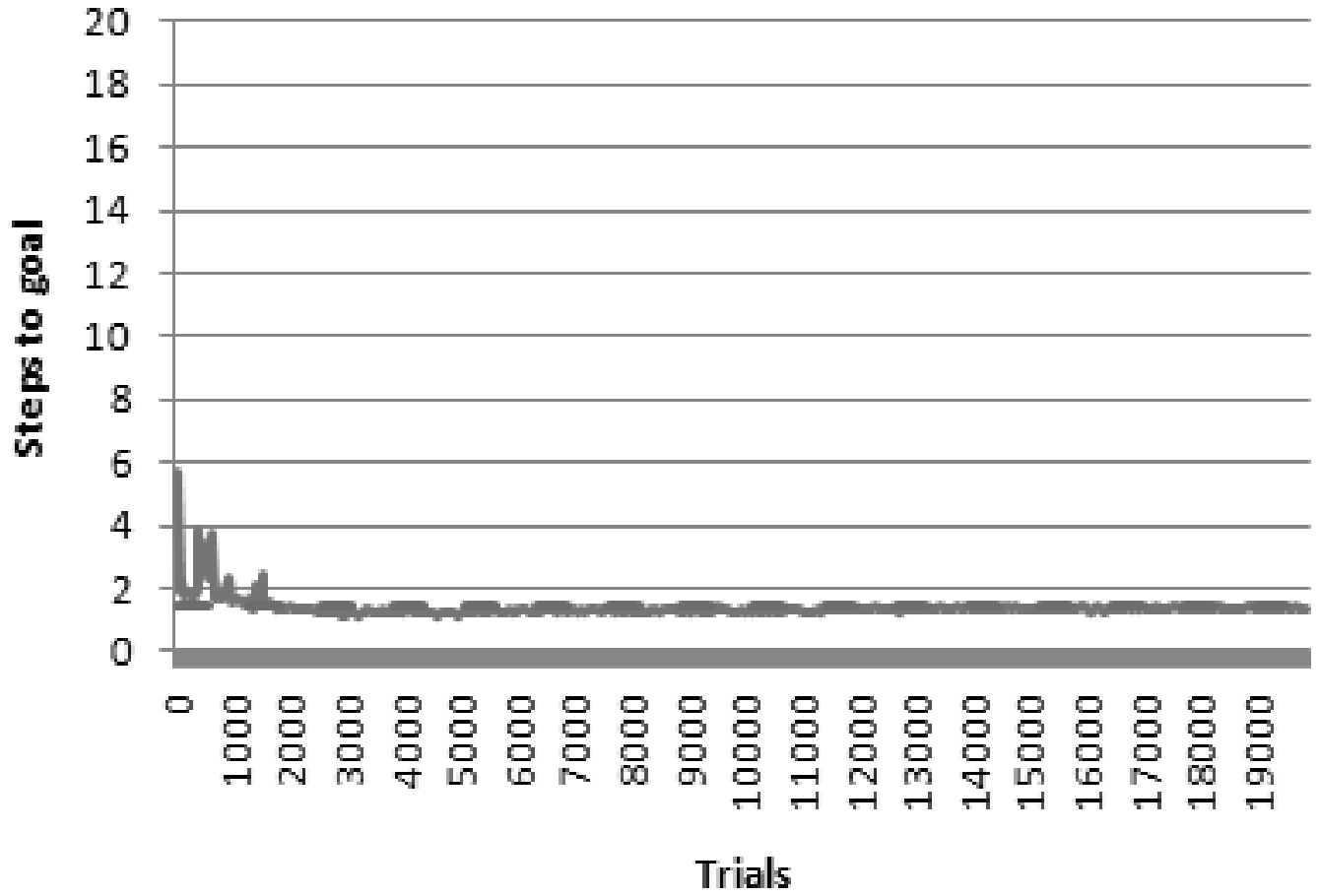,width=8cm,height=4cm}}
\subfloat[]{ \psfig{file=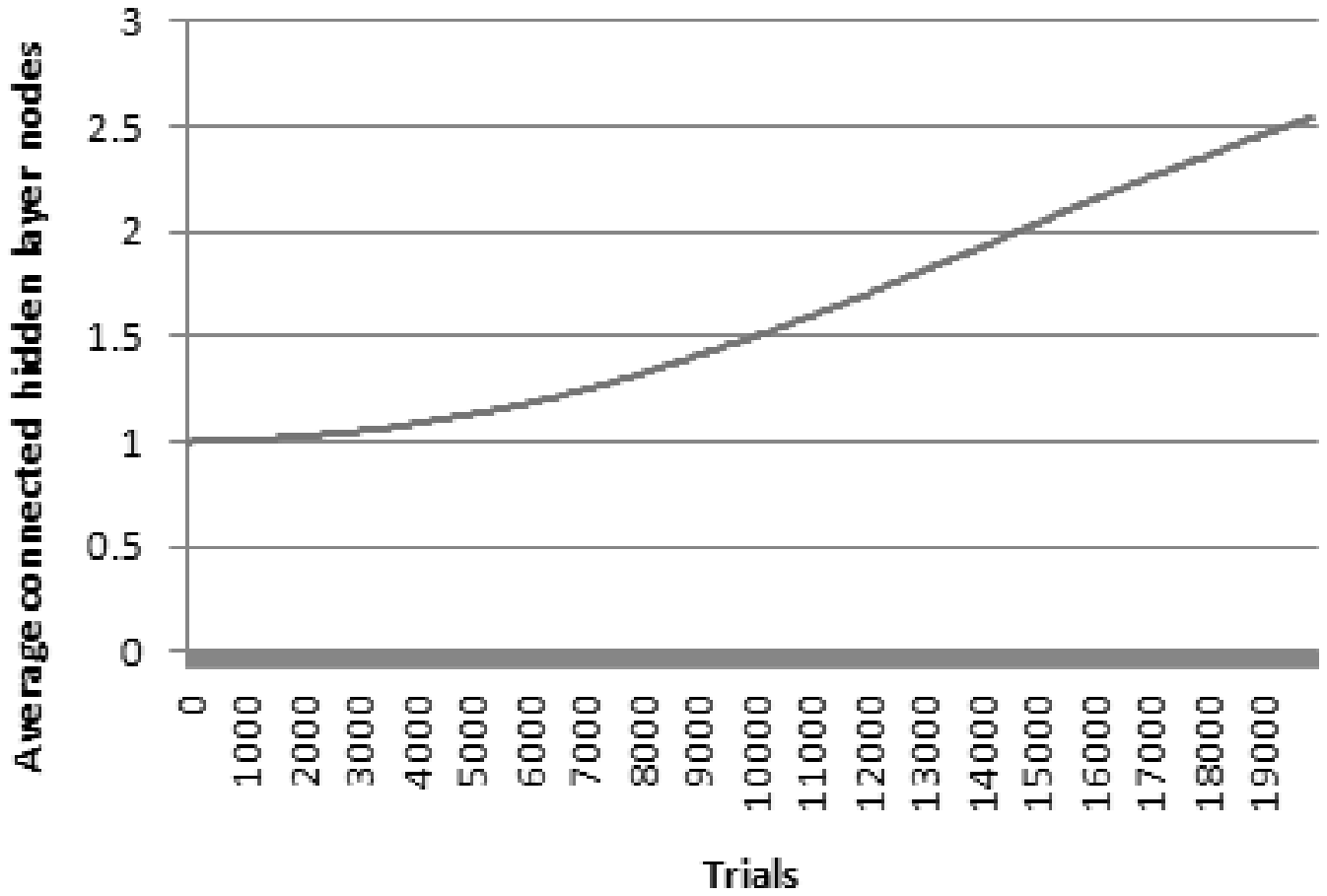,width=8cm,height=4cm}}\\
\subfloat[]{ \psfig{file=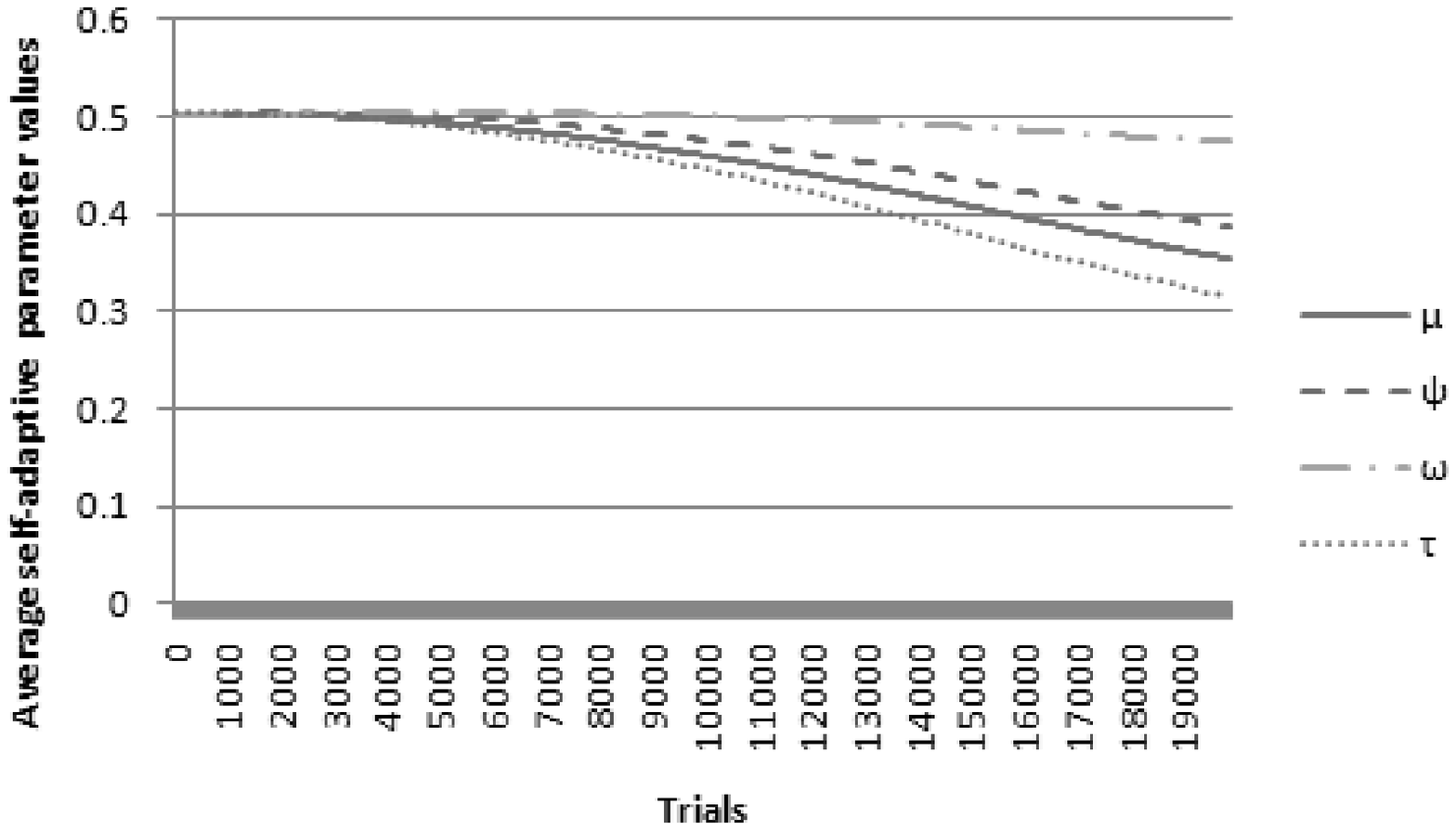,width=8cm,height=4cm}}
\subfloat[]{ \psfig{file=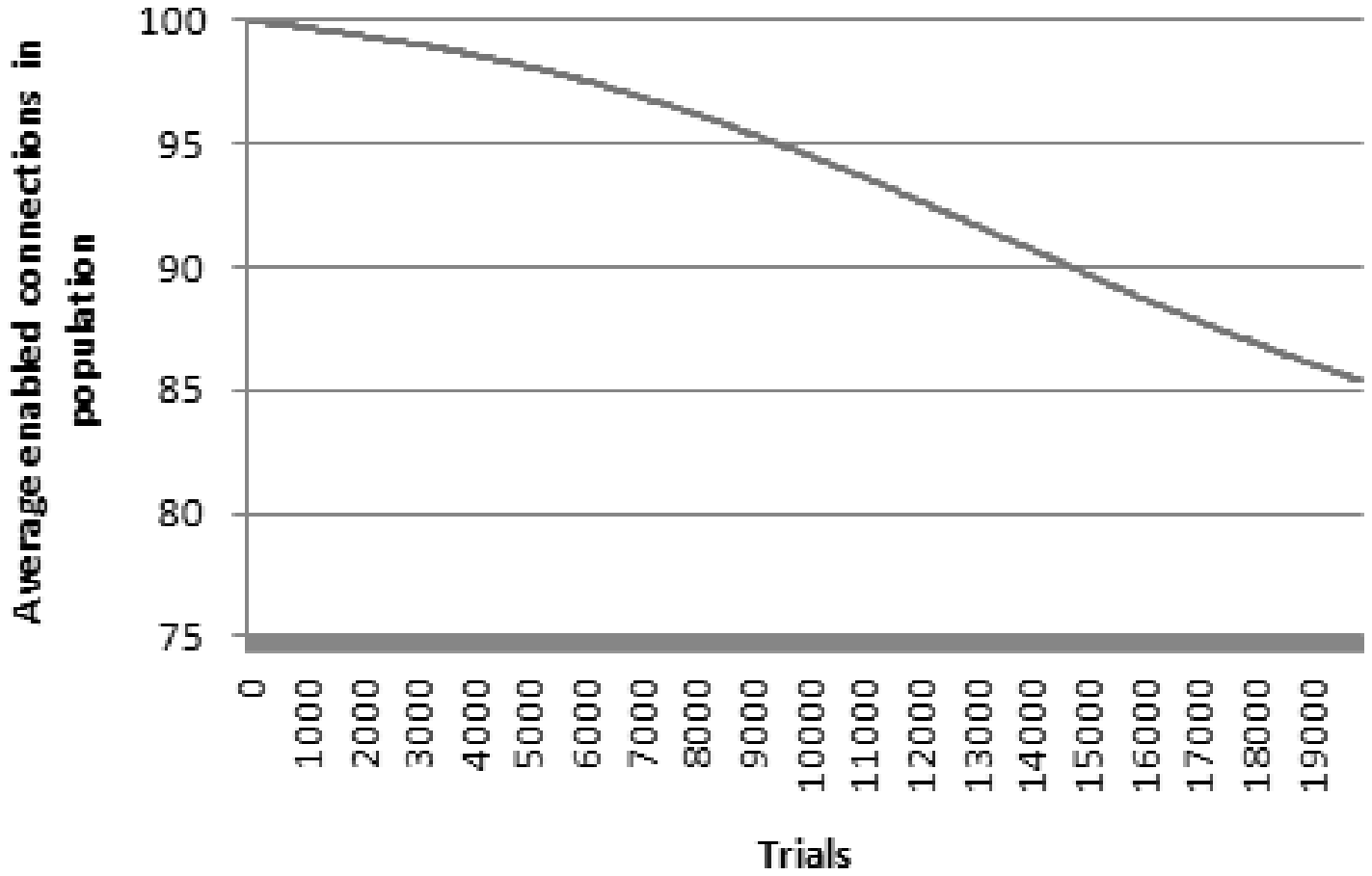,width=8cm,height=4cm}}

\end{center}
\caption[]{Continuous grid world with step size 0.005 (a) Steps to goal, (b) average connected hidden layer nodes, (c) average self-adaptive parameter values, (d) average enabled connections in spiking TCS N-XCSF}
\label{Spike-TCS-0005}
\end{figure*}

We increased the timeout value from 20 to 200, giving an optimal average steps-to-goal value of 1.5.  To prove the scalability of the system, all other LCS parameters are identical.  Results of comparisons between the spiking TCS-enabled systems within the step size 0.05 and 0.005 environments are presented in Table~\ref{spike-TCS-005-0005}, and Figs.~\ref{Spike-TCS} and~\ref{Spike-TCS-0005}.  Both systems solve the task optimally.  However, the performance of the system with step size 0.005 is statistically better (p=2.42$\times$10$^{-6}$).  A possible explanation for this is that, as the average number of discrete movements an agent is required to make is much greater than in the step size 0.05 case, the SNN networks have more opportunity to use the temporal information in this semi-MDP, resulting in a performance difference.  Fig.~\ref{Spike-TCS-0005}(a) shows the performance of the system in the step size 0.005 environment.  Commencing from an average steps-to-goal value of 5.8, the system initially produces fluctuating results until around 2500 trials; stability is thereafter attained.  Final solutions show that more macroclassifiers are required in the smaller step size environment in an almost statistically significant manner, p=0.03.  This indicates that more network variety is required in a more granular environment.  Both step sizes produce solutions with a similar number of connected nodes, although the smaller step size does induce a slight growth in node-wise network complexity (2.4 connected hidden nodes on average in Fig.~\ref{Spike-TCS}(b), contrasting with 2.5 nodes on average in Fig.~\ref{Spike-TCS-0005}(b)); p=0.15.  Again, the curvatures are similar, but  final values differ slightly.  A possible explanation for the increased number of hidden nodes is that, as the step size 0.005 environment potentially provides more latent temporal information per trial (as more discrete steps are required on average), more hidden nodes (temporal processing units) are evolved by the networks to usefully process this latent information.

\begin{table}[t!]
\begin{center}
\caption[]{Detailing t-test results in TCS spiking  N-XCSF in the continuous grid world with step sizes 0.05 and 0.005}
\label{spike-TCS-005-0005}
\begin{tabular}{|l|l|l|l|} 
\hline Metric & Step size & Average & P-value \\ 
\hline Stability & 0.05 & 5637.5 &  \\ 
  & 0.005 & 1244.13 & 2.42$\times$10$^{-6}$ \\ 
\hline Mutation & 0.05 & 0.25 &  \\ 
 & 0.005 & 0.35 & 0.01 \\ 
\hline Neurons & 0.05& 2.40 & \\
 & 0.005& 2.55&0.15 \\
\hline Connectivity &0.05& 86.13 & \\
& 0.005 &85.09 & 0.36 \\
\hline Macroclassifiers&0.05&16552.25 &  \\
 & 0.005 &18384.13 &0.03  \\
\hline 
\end{tabular}
\end{center}
\end{table}

Immediately obvious were similar parametric trends appearing in both sets of graphs.  For example, the line curvatures for self-adaptive parameters described in Figs.~\ref{Spike-TCS}(c) and~\ref{Spike-TCS-0005}(c) are similar in shape, and follow an identical descending order ($\omega$, $\mu$, $\psi$, $\tau$) – the only difference being that the parameter values in Fig.~\ref{Spike-TCS}(c) are slightly lower in general (reaching values of 0.44, 0.29, 0.25 and 0.22 respectively with step size 0.05, compared with 0.48, 0.39, 0.36 and 0.31 in the step size 0.005 environment).  Differences between the self-adaptive parameter values in Fig.~\ref{Spike-TCS-0005}(c) can only be seen after approximately 10000 trials.  Table~\ref{spike-TCS-005-0005} reveals that the final average self-adaptive mutation values vary statistically significantly between the different step size environments, p=0.01.   Fig.~\ref{Spike-TCS-0005}(d) shows an average network connectivity of approximately 86\%, whereas Fig.~\ref{Spike-TCS}(d) shows 81\%, p=0.36.  The number of macroclassifiers required does not vary significantly between the trials (p=0.03).  Despite similar trends, differing values again highlight the self-adaptive nature of the learning process, which alters depending on the environment the system is presented with.  

As in Sect. 6, we compared performance to that of a baseline tabular Q-learner with step size 0.005 and discount rate $\gamma = 0.99$.   We performed the optimal performance-giving discretisation as given in equation 8 to discretise the continuous space into a 201 $\times$ 201 grid based on this new step size.  With a  steps-to-goal value always $>$400 (optimal 181.4, results not shown),  Q-learning was unable to solve this more challenging environment within the allotted timeframe.  The primary reason for this failure can be explained as too fine a sampling granularity within the environment - the system could not generate an accurate payoff map of the environment as there were too many discount steps in an average trial to successfully utilise Q-learning within a reasonable timeframe.  Hence we show the effectiveness of the temporal nature of TCS – the ability harness the temporal sensitivity of the SNN classifiers to deal with long action chains (continuous actions) that potentially consist of multiple discrete actions, with no need to pre-discretise the continuous state space, in the presence of sensory noise.  

It should also be noted that these experiments were also carried out on the spiking non-TCS and MLP TCS versions of our system (results not shown); neither could find the optimal discretisation as evidenced in the spiking TCS version,  demonstrating the ability of SNNs to harness temporal information and highlighting the suitability of TCS-style functionality for a simulated robotics environment.

\subsection{Mountain-Car}

To demonstrate the general learning ability of spiking TCS, we also tested on the mountain-car problem \citep{sutton-scc}, in which
a car must be guided out of a one-dimensional valley.  Reaching the goal state is non-trivial as, in some cases, the car must move away from the goal state to attain enough momentum to climb out of the valley.  State variables were {\em position} [-1.2, 0.6], and {\em velocity} [-0.07, 0.07].  Three actions were available: forward (increase velocity), backward (decrease velocity), and no movement.  For a complete algorithmic description see \citep{sutton-scc}.  As there were only three actions available, SNN outputs were distretised into actions as: forward = {\em high, high}, backward = {\em low, low}, and no movement = {\em high, low} or {\em low, high}.  Each experiment consisted of 5000 trials, 2500 explore and 2500 exploit.   The agent was initally randomly placed in the environment (except in the goal state), given a random velocity,  and had to reach the goal state (where the cars position is $>0.5$) in the fewest  possible steps (optimal steps-to-goal =1).    Population size $N$=1000; TCS parameters were $\varphi$=0.45, $\rho$=0.005, $timeout$=200.  All other parameters were identical to those in Sect. 7.

\begin{figure*}[t!]
\begin{center}

\subfloat[]{ \psfig{file=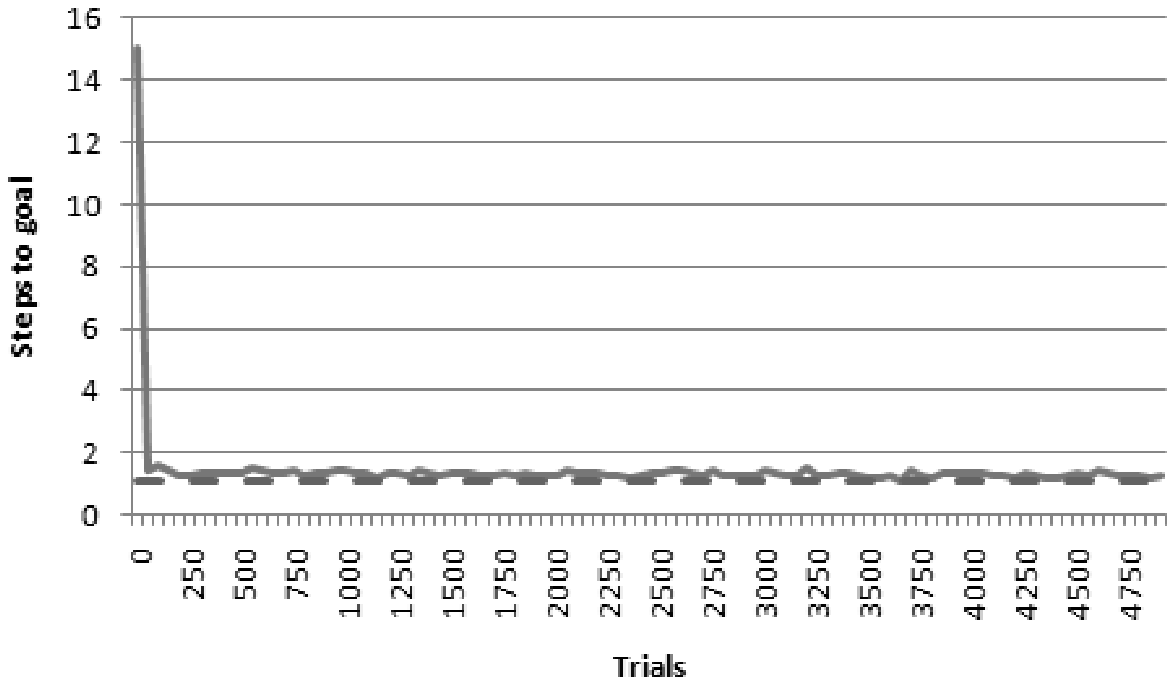,width=8cm,height=4cm}}
\subfloat[]{ \psfig{file=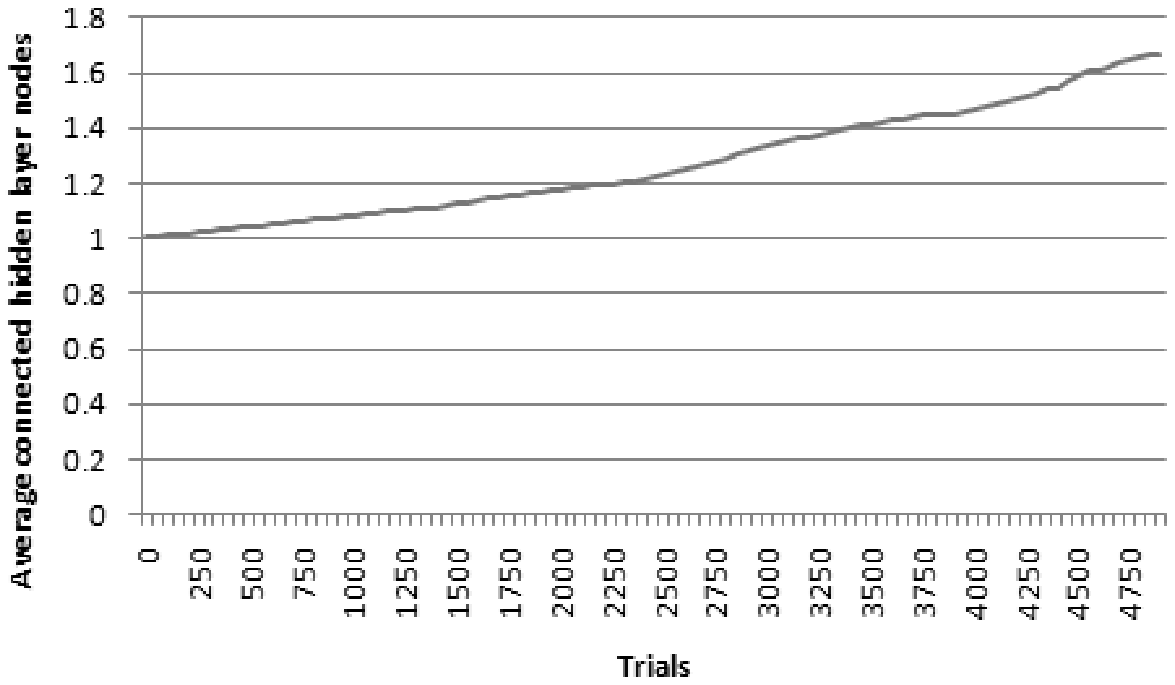,width=8cm,height=4cm}}\\
\subfloat[]{ \psfig{file=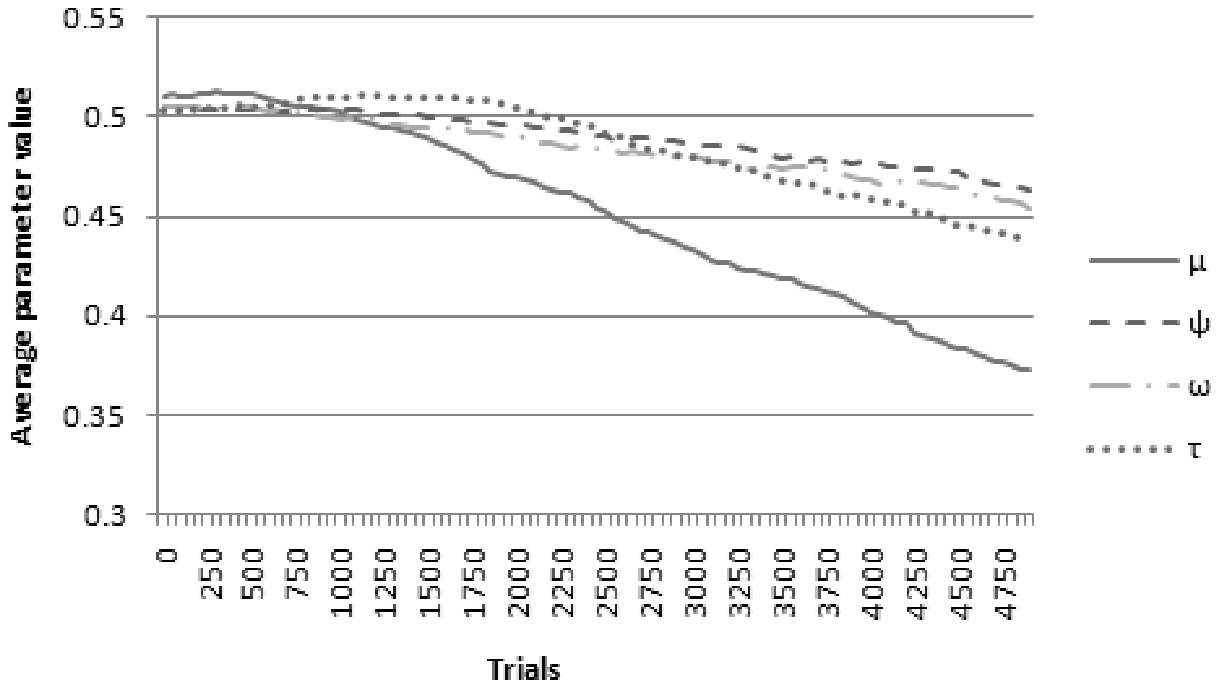,width=8cm,height=4cm}}
\subfloat[]{ \psfig{file=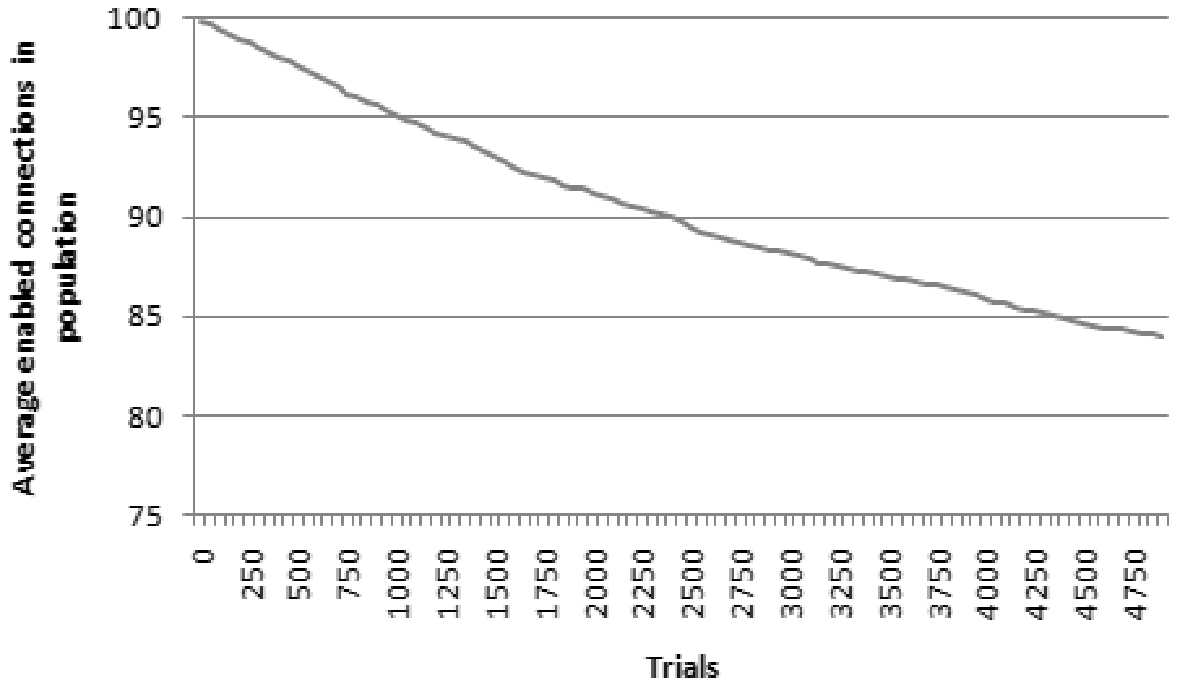,width=8cm,height=4cm}}

\end{center}
\caption[]{Mountain-car (a) Steps to goal, (b) average connected hidden layer nodes, (c) average self-adaptive parameter values, (d) average enabled connections in spiking N-XCSF}
\label{mcar}
\end{figure*}


Fig.~\ref{mcar}(a) shows attainment of optimum performance within 100 trials, which appears competitive to XCSF with a tile coding (e.g.  \citep{journals/ras/Tham95}) scheme ~\citep{lanzi-tile}.  It should be noted that ~\citep{lanzi-tile} allows only two actions.  Average neurons per classifier (Fig.~\ref{mcar}(b) gradually increases to a final value of ~1.65.  Self-adaptive parameters decline from their initial values (Fig. ~\ref{mcar}(c)).  Final network connectivity (Fig.~\ref{mcar}(d)) steadily declines to a final average value of 84.5\%.  Action set analysis shows that TCS allows the agent to reach the goal state from any intial position/velocity combination by forming only one match set.  Actions are altered from forward to backward as the agent builds momentum; this is achieved by the networks recalculating their actions based on sensory input to alter the system action.  The TCS reinforcement formula allows for the generation of shortest-length action chains within these macro actions within 480 trials.

\section {Moving to Robotics Problems}

At the beginning of Sect. 7, justification for the use of a continuous test environment summarised as a desire to more accurately replicate the types of situations a real robot might encounter.  Here, we begin with explanation for the transition from the continuous maze to a physical robotics simulation.  Although a standard RL test scenario, the continuous grid world is an environment lacking in any {\em real-world} complexity,  selected mainly to allow a direct comparison to tabular Q-learning.  Evaluation of our system on a robot simulation suite is therefore a logical step forwards.  An overview of current robot simulation platforms can be found in \citep{CraigheadMBG07}.  Our chosen development simulator is the Webots robotics platform \citep{webots04}, a test bed chosen due to its popularity in the research community.

\subsection{LCS Robotics research}

The first LCS to successfully tackle the world of real robot control was the work of \cite{journals/ai/DorigoC94}.  The authors used a modified version of Hollands’ classifier system \citep{Holland1978}, to create a hierarchical LCS in which lower-level LCSs can learn simple behaviours, which high-level LCSs then coordinate to generate complex actions.  MONALYSA \citep{Donnart1996c} was a hierarchicial LCS in which the hierarchy itself could be dynamically reconfigured.  It was experimentally demonstrated that the ability to hierarchically decompose the required behaviour into sub-behaviours produced better performance, firstly in a simulated environment and later on a physical robot.  

\cite{Bonarini1998a} presented a fuzzy classifier system that created sets of fuzzy niches which guided the behaviour of the agent, and employed a delayed reinforcement attribution scheme similar to Q-learning.  The authors reported swift learning compared to the traditional ternary classifier condition representation.  \cite{journals/isci/BonariniT01} then extended this system for cooperation amongst a swarm of agents who could explicitly pass messages between themselves within a limited range.  Another fuzzy classifier system is introduced by \cite{Pipe:2002:FRE} for implementation on a robotics platform.  The system was tested on non-trivial maze environments, which were navigated by a physical robot;  compact rule sets were reported as a benefit of fuzzy classifier representation. 

Latent learning was realized in a classifier system by \cite{stolzmann}, providing a degree of premonition.  It was demonstrated that his LCS could build chains of classifiers without waiting for subsequent environmental inputs by creating its own internal representation of the environment.  

\cite{kata-yam} presented a manual approach for inducing certain behavioural patterns as a bootstrapping technique for learning in a real robot.  Reinforcement learning was used as normal to decide on the ``best'' actions to take from the initially generated set.  The operator could also decide to take control in which case system created rules that covered the actions that the operator took,  although this is obviously supervised learning.  \cite{conf/ecal/WebbHRL03} reported mixed success in XCS control over a simulated Khepera in Partially Observable Markov Decision Process (POMDP) maze environments. They attempted to bypass the ``aliasing states'' problem by augmenting each classifier with an internal state register and internal action, which were used to differentiate between aliasing states.  The LCS made use of temporal actions with TCS-like functionality.  A simple LCS is presented by \cite{conf/cec/CazangiZF03}, who evolved robot controllers for goal location and object avoidance tasks in unknown environments.  The authors reported no significant degradation of performance when switching from simulation to a physical agent and also demonstrated that successful controllers can be evolved entirely on the physical robot.  More recently, \cite{conf/gecco/ButzH08} demonstrated an XCSF-derivative to control a robot arm using a real interval representation.   Recent research by \cite{conf/iwcls/MoioliVZ07} compared attempts at T-maze navigation in both simulated and real robotics tasks, and presented results that reinforce the idea that TCS is a valid method for dealing with long-action chains that are present in many robotic environments.  Similarly to \cite{conf/ecal/WebbHRL03}, the authors  added a memory register and tested the system on a non-reactive robotics task.  Again, high performance and the ability to disambiguate perceptually aliased states is reported.   It should be noted that TCS-based systems have previously been used explicitly for real robot navigation, by \cite{conf/cec/StudleyB05} and \cite{journals/alife/HurstB06}.

\subsection{Experimentation}
\begin{figure}[t]
\begin{center}

\subfloat[]{ \psfig{file=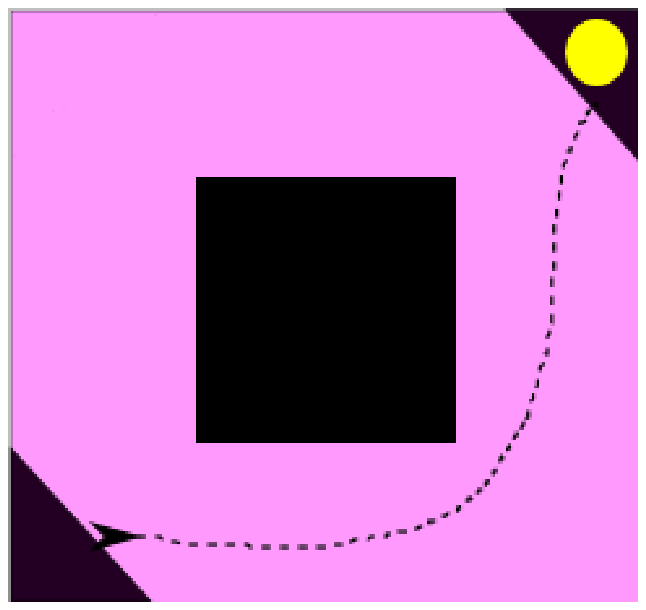,width=3.5cm, height=3.5cm}}
\subfloat[]{ \psfig{file=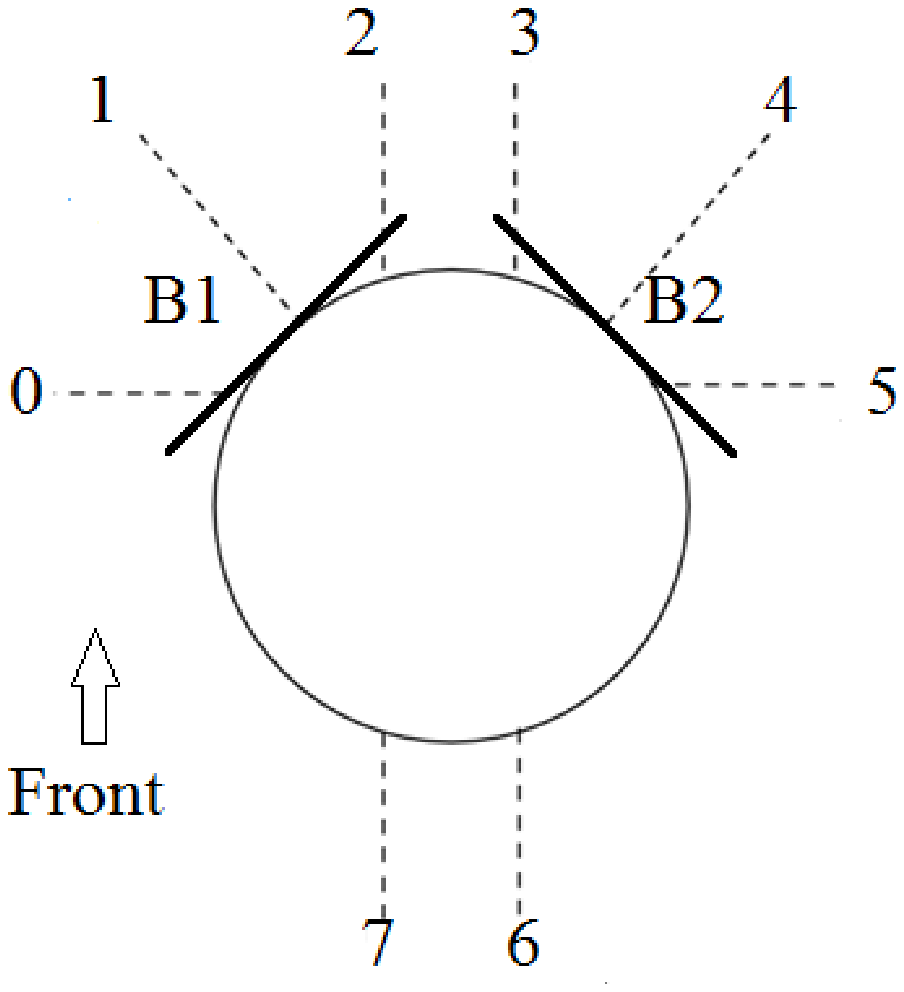,width=3.5cm, height=3.5cm}}
\\

\end{center}
\caption[]{(a)The test environment.  The agent begins in the lower-left and must reach a light source (circle) in the upper-right, circumnavigating the central obstacle.  An example agent path is shown (dotted line).(b)Khepera sensory arrangement.  3 light sensors and 3 IR sensors share positions 0, 2 and 5.  Two bump sensors, B1 and B2, are shown attached at 45 degree angles to the front-left and front-right of the robot.}
\label{khep-sens}
\end{figure}

This final set of experiments demonstrates the performance of the spiking TCS N-XCSF in a robotics environment (Fig.~\ref{khep-sens}(a)).  The agent, a simulated Khepera II robot,  was initially randomly located within a walled arena which it could not leave with coordinates ranging from [-1,1] in both $x$ and $y$ directions (all units are in metres). A light source was placed at the top-right hand corner of the arena ($x$=1, $y$=1, $z$=1), which the agent must approach in order to receive reward.  The light source was modelled on a 15W bulb with realistic attenuation values.  Adding to the complexity of the environment, a three-dimensional box was placed centrally in the arena (with vertices on ``ground level'' ($z$=0.0 ) at ($x$=-0.4, $y$=-0.4), (-0.4, 0.4), (0.4,0.4), and (0.4, -0.4), and raised to a height of $z$=0.15).  When the agent reached the reward zone (where $x+y>$1.6), an immediate reward of 1000 was returned and the next trial begun.  All other movements give an immediate reward of 0.  The reward boundary of 1.6 was chosen as the result of calibration experiments, which revealed the agent experienced excessive levels of sensory aliasing when approaching the light.  All IR and light sensor lookup tables were modelled on actual sensor performance from the real Khepera robot.

We altered the environmental representation used by the agent to calculate action and predict payoff.  Contrasting our previous approach of using the agent’s noisy ($x$,$y$) location in the environment as the input state $s_t$, we used readings taken from the agent’s light and distance sensors to model the current environmental state.  The agent, a cylinder-shaped robot with height = 0.03 and diameter 0.07,   was equipped with with 8 light sensors and 8 IR distance sensors (see Fig.~\ref{khep-sens}(b)), plus two bump sensors, offset to the left and right of the front of the agent.   At each step, the agent sampled its light and IR sensors, whose scaled values ranged [0,1].  These values then comprised the input state for the current step.  After \citep{journals/alife/HurstB06}, six sensors were used to comprise the input state, three IR and three light sensors at positions 0, 2 and 5 (see Fig.~\ref{khep-sens}(b)).  This extended input state was thereafter used to calculate [M] membership/actions and compute prediction as normal.

In this experiment we employed our spiking TCS N-XCSF to solve a more complex version of the grid environment introduced in Sect. 6.1.  This environment was much more challenging than any we have used previously.  For example, the agent’s wheels could slip, sensory noise was more pervasive and more accurately modelled, and the environmental input was three times larger than that used in the continuous grid world experiments.  The inclusion of an obstacle further increased this complexity.  At the start of each trial the agent’s initial random position was further constrained so that the obstacle is initially always between the agent and the light source.  We constrained the initial starting position the lower left-hand corner of the environment, where the  inequality ($x + y < -1.5$) is satisfied.

To demonstrate the problem-independence of the system, parameter settings were largely unchanged from those in Sect. 8, with  exceptions.  First, {\em  N} was reduced to 3000.  Due to time and processing constraints, each experiment was limited to 500 trials.   Secondly, each classifier was initially seeded with 6 hidden layer nodes  to offset the lack of trials per experiment.  Seeding the networks with one neuron would make it extremely difficult for the classifier to immediately and usefully discriminate certain state elements as it is assumed that providing 6 state elements instead of the normal 2 would require the networks to make more complex partitions in state space.  To reduce disruption, self-adaptive parameters were initially constrained to (0$<(\mu/\psi/\tau) \leq$0.02), with (0$<\omega \leq$1) as normal.  Finally, the six hidden layer nodes had each initial connection enabled with 50\% probability.  This is intended to increase inter-network variation early in the experiment as certain state variables will initially have no effect on the output action, thus increasing behavioural diversity.  Usually, connection selection performed this operation gradually throughout the 20,000 trials.  All of these modifications were implementated to aid expediency; as we completed only 500 trials, the new parameterisation allowed the networks to perform useful fucntions immediately.  Due to the nature of construcitivism, it is assumed that the system would eventually be able to learn an optimal policy for the given task given enough time when set to its default parameters.

Movement values and sensory update delays were constrained by accurate modelling of physical Khepera agent.  It should be noted that in the earlier experiments, agent orientation was irrelevent; here orientation was preserved through movements and the agent was able to turn continuously to explore the environment.  Three actions were possible: {\em forward}, and continuous turns to both the {\em left} and {\em right} (caused by halving the left/right motor outputs respectively).  As the agent initially explored the environment, it was likely to bump into obstacles.  If either bumper was activated, an interrupt was sent causing the agent to reverse 10cm and form a new [M] (after \citep{journals/alife/HurstB06}). 

\begin{figure*}[ht!]
\begin{center}

\subfloat[]{ \psfig{file=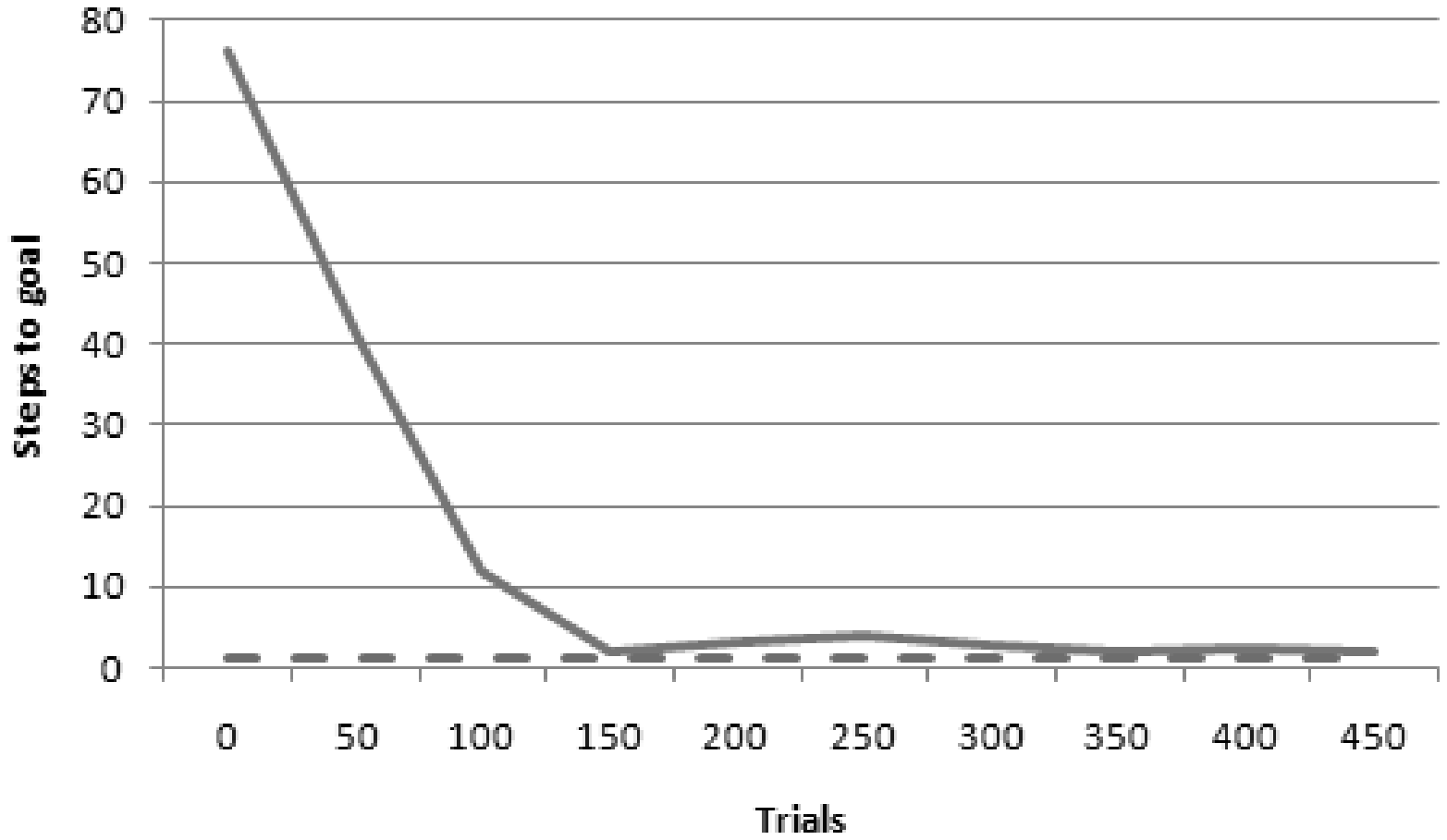,width=8cm,height=4cm}}
\subfloat[]{ \psfig{file=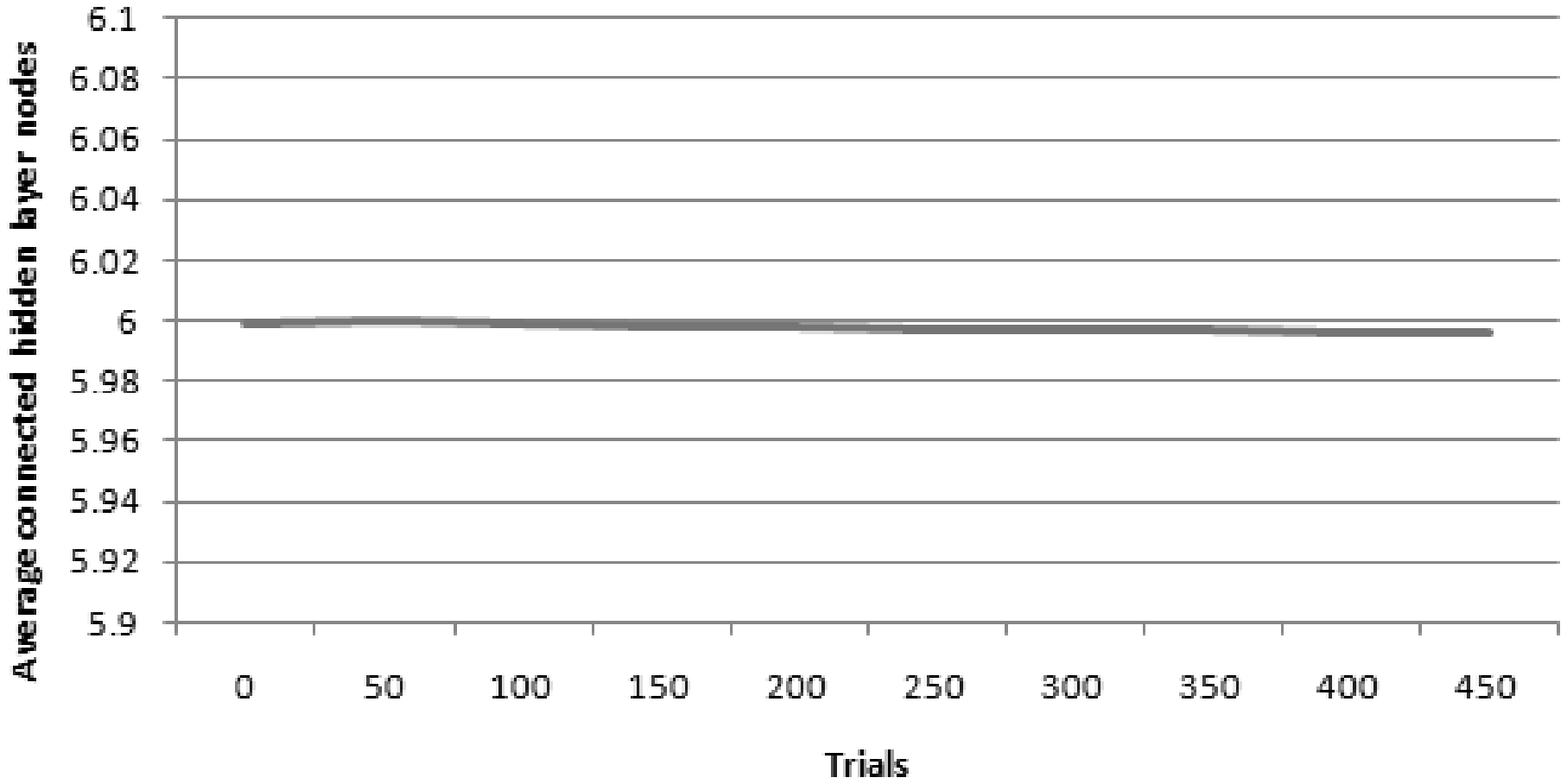,width=8cm,height=4cm}}\\
\subfloat[]{ \psfig{file=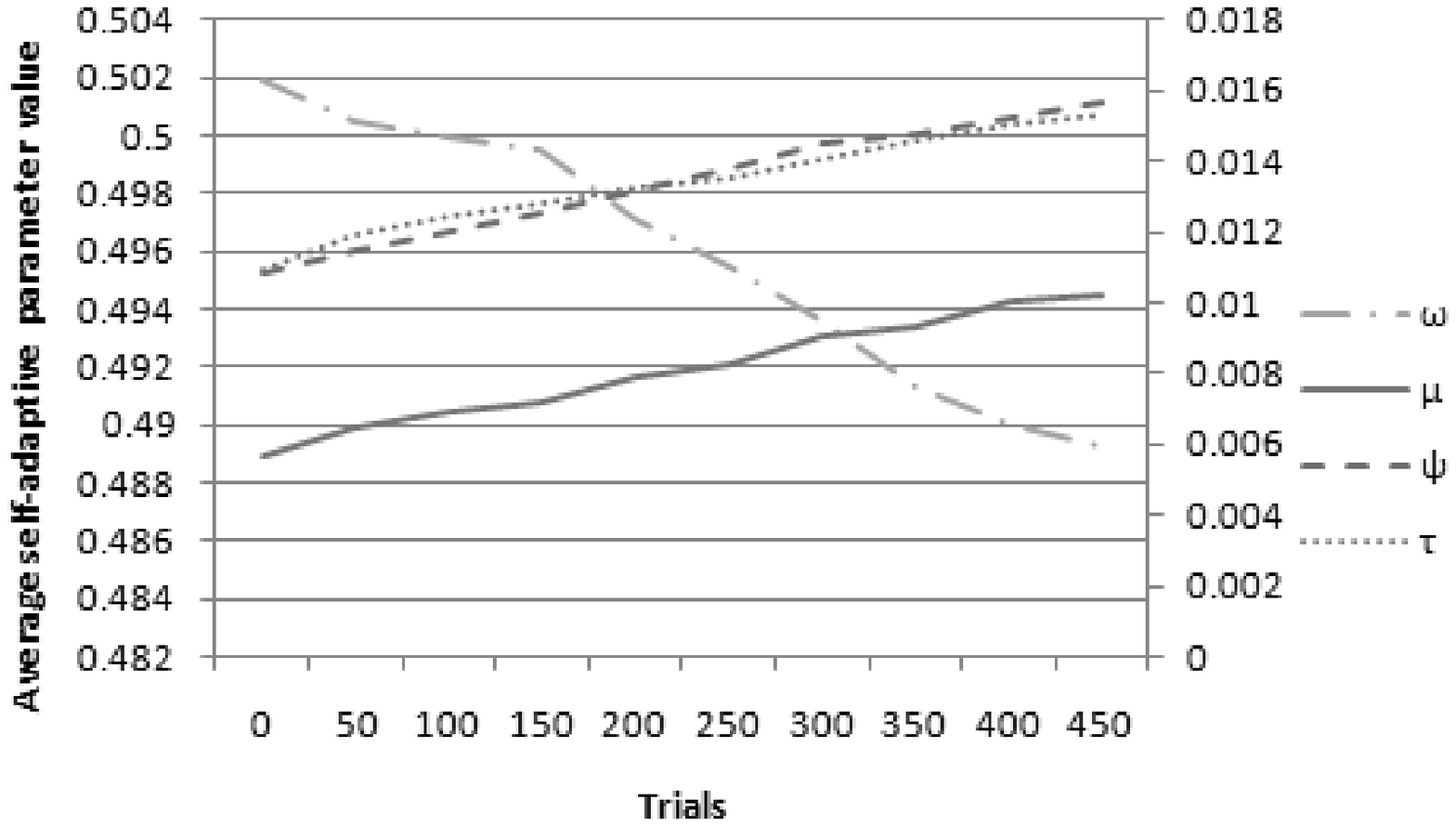,width=8cm,height=4cm}}
\subfloat[]{ \psfig{file=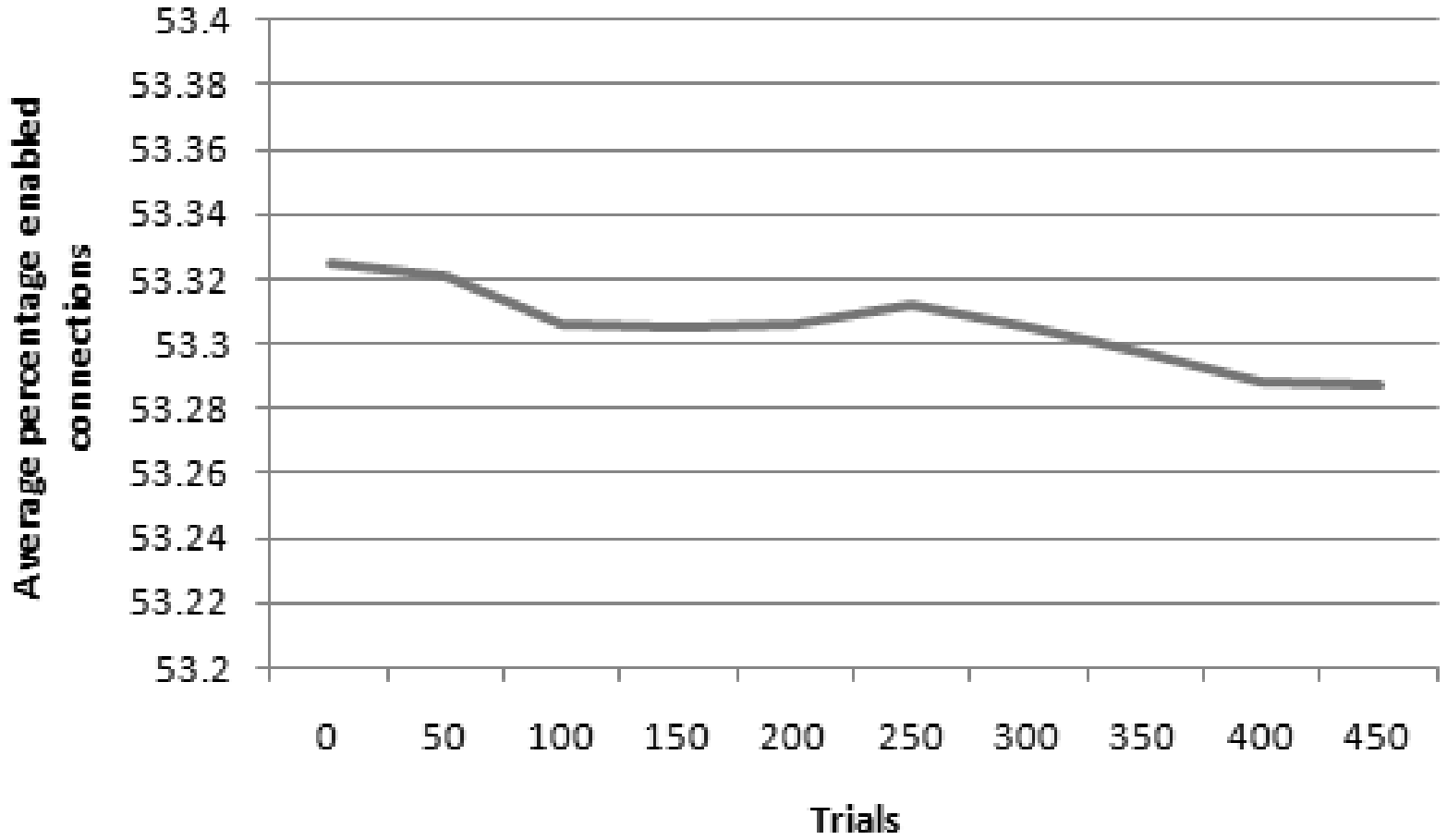,width=8cm,height=4cm}}

\end{center}
\caption[]{(a) Steps-to-goal (b) average connected hidden layer nodes (c) average self-adaptive parameter values ($\mu/\psi/\tau$ plotted on the right axis) (d) average enabled connections for TCS N-XCSF in the Webots experiment}
\label{webots-sim-grid}
\end{figure*}

Fig.~\ref{webots-sim-grid}(a) shows the steps-to-goal values attained.  Starting from 75 steps-to-goal, the system shows swift attainment of near optimal performance in terms of number of match set formations per trial after 300 trials, especially considering the levels of sensory noise.  It is demonstrated that the system could, in most cases, evolve networks flexible enough to perform segregations in action space without reforming [M] via the ability of the networks to recalculate actions within an action set due to differences in received states.  

An initial exploratory phase of action selection involved many collisions, which were progressively penalized as they caused more [M] formations per trial.  The responsible classifiers were progressively usurped by classifiers which retained the same initial [M] by avoiding collisions.  Action alteration once [M] was formed was usually due to IR sensors activating and perturbing network performance to turn the agent away from an obstacle or towards the light source, although the agent was also observed to display the capability to alter action based on light sensor input alone.  During a trial, action calculation was seen to be altered in two ways; networks either evolved to activate the ``don’t match'' node so that classifiers advocating certain actions were dropped from [A] at the correct time, or state inputs caused the action advocated by the majority of the classifiers in the current [A] to transition between two states, causing a turn.  Networks were observed to use the latter method more frequently than the former (68\% of action switches occurred using the latter method), an evolved behaviour to keep more classifiers in [A] and hence retain more classifier variety in the action set.  It was interesting to note that once the correct amount of turn was applied to the agent (e.g. in response to IR sensors highly activating, then deactivating once the obstacle was avoided), the action set was observed to re-calculate the ``correct''  action immediately;  both methods outlined above were observed to be used by the system.  Varying amounts of turn were observed (as more frequent {\em turn} actions interspersed within a majority of {\em forward} actions) as the agent approached an obstacle.

The number of hidden layer nodes are shown in Fig.~\ref{webots-sim-grid}(b), self-adaptive parameters shown in Fig.~\ref{webots-sim-grid}(c), and connectivity shown in Fig.~\ref{webots-sim-grid}(d).  All graphs show that the initial parameter values are mainly unaltered, due to the reduction in the number of trials.   Performance was seen to compare favourably to other TCS robotics experiments, (e.g. \citep{journals/alife/HurstB06}).  As mentioned previously, by seeding each classifier with 6 hidden layer neurons, and allowing any connection in the network to be initially disabled with 50\% probability, we effectively ``jumped'' the start of constructive neuro-evolution, allowing the networks sufficient topological disparity and neural complexity to begin solving this more complex problem immediately.

Numerous differences existed between the simulation and traditional grid world experiments.  Firstly, variations in terms of state space were seen to present a more difficult experiment for the simulated robot to navigate; the value function in the previous continuous grid world experiments increased smoothly as the agent neared the goal state.  The value function also mapped approximately to the sensory readings perceived by the agent so that higher values of $x$ and $y$ gave, in general, higher payoffs. Conversely, in simulation the robot had to deal with less uniform state space due to the short range (0.01) of the IR sensors, which could within a short amount of time drastically alter the state input to the networks;  hence the payoff map did not correlate with its' state as simply.  The environmental representation was more complex; the light sensors were easily saturated, and the environmental was mainly well-illuminated; a necessary implementation restriction due to using Webots to ensure that collision detection occurs.  It was noted that there was sparse use of high ($>$0.5) light sensor values that correspond to very dark regions of the environment.  Also prevalent was a degree of sensory noise surpassing the [+/-5\%] used in the continuous grid world environment; IR sensors were [+/-5\%] at the extremes of their range, light sensors [+/-10\%].

\section{Conclusions}
In this paper we have shown that a self-adaptive neural LCS employing constructivism can perform optimally in a more complex and noisy version of a standard continuous environment, and a continuous noisy robotic simulation.  As the networks can calculate their actions, they have the capability to carry out the correct action in different areas of the problem space, even if that action required is different.   Further, using the prediction computation of XCSF, we have observed that one network can accurately predict payoff in several spatially disparate regions of the problem space, even when the payoff values are different.  

The results of adding TCS showed that when the system was presented with an environment which allowed it to harness its temporal capabilities, it performed in excess of environments where a temporal element does not exist, in other words the system is capable of processing underlying temporal information in a problem.  The main strength of the TCS approach was the generation of high-level continuous actions from simple discrete actions, which allowed the agent to traverse environments requiring long action-chains.  As the LCS can search the space of continuous actions, removing the need to predefine such actions.  By adding SNN classifiers, the action advocated at a given timestep could be recalculated based on sensory input, increasing generalisation.

Results presented in this paper are intended to reinforce the view that the use of self-adaptation and constructivism are a means to achieving increased levels of parameter and problem independence.  XCS and XCSF are notoriously heavily parameterised systems; self-adaptation removes the need to set several parameters and constructivism allows the LCS to adapt to the complexity of the problem it is presented with.  In the case of the transition to Webots, a much more complex environmental representation is implemented without significant parameter change.  The transition to Webots demonstrated the power of the system to create hetergeneous continuous movement sequences from a single match set formation, despite levels of sensory noise in simulation being in excess of the emulated noise used in the continuous maze environment.

\bibliographystyle{spbasic}
\bibliography{xcsf-sc}   

\begin{thebibliography}{69}
\providecommand{\natexlab}[1]{#1}
\providecommand{\url}[1]{{#1}}
\providecommand{\urlprefix}{URL }
\expandafter\ifx\csname urlstyle\endcsname\relax
  \providecommand{\doi}[1]{DOI~\discretionary{}{}{}#1}\else
  \providecommand{\doi}{DOI~\discretionary{}{}{}\begingroup
  \urlstyle{rm}\Url}\fi
\providecommand{\eprint}[2][]{\url{#2}}

\bibitem[{Ahluwalia and Bull(1999)}]{Ahluwalia1999}
Ahluwalia M, Bull L (1999) A genetic programming-based classifier system. In:
  Banzhaf W, Daida J, Eiben AE, Garzon MH, Honavar V, Jakiela M, Smith RE (eds)
  Proceedings of the Genetic and Evolutionary Computation Conference
  ({GECCO}-99), Morgan Kaufmann, pp 11--18

\bibitem[{Bacardit et~al(2007)Bacardit, Stout, Hirst, Sastry, Llor{\`a}, and
  Krasnogor}]{BacarditSHSLK07}
Bacardit J, Stout M, Hirst JD, Sastry K, Llor{\`a} X, Krasnogor N (2007)
  Automated alphabet reduction method with evolutionary algorithms for protein
  structure prediction. In: Lipson H (ed) Genetic and Evolutionary Computation
  Conference, {GECCO} 2007, Proceedings, London, England, {UK}, July 7-11,
  2007, ACM, pp 346--353

\bibitem[{Bonarini(1998)}]{Bonarini1998a}
Bonarini A (1998) Reinforcement distribution to fuzzy classifiers. In:
  {Proceedings of the IEEE World Congress on Computational Intelligence (WCCI)
  -- Evolutionary Computation}, {IEEE Computer Press}, pp 51--56

\bibitem[{Bonarini and Trianni(2001)}]{journals/isci/BonariniT01}
Bonarini A, Trianni V (2001) Learning fuzzy classifier systems for multi-agent
  coordination. Information Sciences 136(1-4):215--239

\bibitem[{Boyan and Moore(1995)}]{boyan.moore-1995:gener}
Boyan JA, Moore AW (1995) Generalization in reinforcement learning: {S}afely
  approximating the value function. In: Tesauro G, Touretzky DS, Leen TK (eds)
  Advances in Neural Information Processing Systems 7, The MIT Press,
  Cambridge, MA, pp 369--376

\bibitem[{Buhmann(2003)}]{Buhmann}
Buhmann MD (2003) Radial basis functions: theory and implementations, Cambridge
  Monographs on Applied and Computational Mathematics, vol~12. Cambridge
  University Press, Cambridge, UK

\bibitem[{Bull(2002)}]{Bull:2002:UCN}
Bull L (2002) On using constructivism in neural classifier systems. In: Merelo
  J, Adamidis P, Beyer HG, Fernandez-Villacanas JL, Schwefel HP (eds) Parallel
  Problem Solving from Nature - PPSN VII, Springer Verlag, pp 558--567

\bibitem[{Bull(2009)}]{Bull2009a}
Bull L (2009) On dynamical genetic programming: Simple boolean networks in
  learning classifier systems. International Journal of Parallel, Emergent and
  Distributed Systems 24(5):421--442

\bibitem[{Bull and Adamatzky(2007)}]{bull-adam-07}
Bull L, Adamatzky A (2007) A learning classifier system approach to the
  identification of cellular automata. Cellular Automata 2(1):21--38

\bibitem[{Bull and Hurst(2003)}]{bull-hurst-tech03}
Bull L, Hurst J (2003) A neural learning classifier system with self-adaptive
  constructivism. In: Proceedings of the IEEE Congress on Evolutionary
  Computation, IEEE Press, pp 991--997

\bibitem[{Bull and O'Hara(2002)}]{BullO02}
Bull L, O'Hara T (2002) Accuracy-based neuro and neuro-fuzzy classifier
  systems. In: Langdon WB, Cant{\'u}-Paz E, Mathias KE, Roy R, Davis D, Poli R,
  Balakrishnan K, Honavar V, Rudolph G, Wegener J, Bull L, Potter MA, Schultz
  AC, Miller JF, Burke EK, Jonoska N (eds) {GECCO} 2002: Proceedings of the
  Genetic and Evolutionary Computation Conference, New York, {USA}, 9-13 July
  2002, Morgan Kaufmann, pp 905--911

\bibitem[{Bull et~al(2000)Bull, Hurst, and Tomlinson}]{Bull2000d}
Bull L, Hurst J, Tomlinson A (2000) Self-adaptive mutation in classifier system
  controllers. In: et~al JAM (ed) From Animals to Animats 6: Proceedings of the
  Sixth International Conference on Simulation of Adaptive Behavior, pp
  460--467

\bibitem[{Butz and Herbort(2008)}]{conf/gecco/ButzH08}
Butz MV, Herbort O (2008) Context-dependent predictions and cognitive arm
  control with {XCSF}. In: Ryan C, Keijzer M (eds) Genetic and Evolutionary
  Computation Conference, {GECCO} 2008, Proceedings, Atlanta, {GA}, {USA}, July
  12-16, 2008, ACM, pp 1357--1364

\bibitem[{Butz et~al(2006)Butz, Lanzi, and Wilson}]{lanzi:2006:hyp}
Butz MV, Lanzi PL, Wilson SW (2006) Hyper-ellipsoidal conditions in xcs:
  rotation, linear approximation, and solution structure. In: GECCO '06:
  Proceedings of the 8th annual conference on Genetic and evolutionary
  computation, ACM Press, New York, NY, USA, pp 1457--1464,
  \doi{http://doi.acm.org/10.1145/1143997.1144237}

\bibitem[{Cazangi et~al(2003)Cazangi, Zuben, and
  Figueiredo}]{conf/cec/CazangiZF03}
Cazangi RR, Zuben FJV, Figueiredo M (2003) A classifier system in real
  applications for robot navigation. In: IEEE Congress on Evolutionary
  Computation, IEEE, pp 574--580

\bibitem[{Cliff and Ross(1994)}]{Cliff1994a}
Cliff D, Ross S (1994) Adding temporary memory to {ZCS}. Adaptive Behavior
  3(2):101--150

\bibitem[{Craighead et~al(2007)Craighead, Murphy, Burke, and
  Goldiez}]{CraigheadMBG07}
Craighead J, Murphy RR, Burke J, Goldiez BF (2007) A survey of commercial \&
  open source unmanned vehicle simulators. In: International Conference of
  Robotics and Automation, IEEE, pp 852--857

\bibitem[{Dolan and Dyer(1987)}]{Dolan1987}
Dolan CP, Dyer MG (1987) Toward the evolution of symbols. In: Grefenstette JJ
  (ed) Genetic Algorithms and their Applications ({ICGA}'87), Lawrence Erlbaum
  Associates, Hillsdale, New Jersey, pp 123--131

\bibitem[{Donnart and Meyer(1996)}]{Donnart1996c}
Donnart JY, Meyer JA (1996) Learning reactive and planning rules in a
  motivationally autonomous animat. IEEE Transactions on Systems, Man and
  Cybernetics - Part B: Cybernetics 26(3):381--395

\bibitem[{Dorigo and Colombetti(1994)}]{journals/ai/DorigoC94}
Dorigo M, Colombetti M (1994) Robot shaping: Developing autonomous agents
  through learning. Artificial Intelligence 71(2):321--370

\bibitem[{Floreano and Mattiussi(2001)}]{Floreano:2001:ESN}
Floreano D, Mattiussi C (2001) Evolution of spiking neural controllers for
  autonomous vision-based robots. Lecture Notes in Computer Science 2217:38--??

\bibitem[{Floreano et~al(2008)Floreano, D\"{u}rr, and
  Mattiussi}]{flor-durr-matt}
Floreano D, D\"{u}rr P, Mattiussi C (2008) Neuroevolution: from architectures
  to learning. Evolutionary Intelligence 1(1):47--62,
  \doi{10.1007/s12065-007-0002-4}

\bibitem[{Gerstner and Kistler(2002)}]{spiking-n-m}
Gerstner W, Kistler W (2002) Spiking Neuron Models - Single Neurons,
  Populations, Plasticity. Cambridge University Press

\bibitem[{Holland(1975)}]{Holland75}
Holland JH (1975) Adaptation in Natural and Artificial Systems. The University
  of Michigan Press, Ann Arbor, Michigan

\bibitem[{Holland(1976)}]{Holland76}
Holland JH (1976) Adaptation. In: Rosen R, Snell F (eds) Progress in
  Theoretical Biology, Academic Press

\bibitem[{Holland and Reitman(1978)}]{Holland1978}
Holland JH, Reitman JS (1978) Cognitive systems based on adaptive algorithms.
  In: Waterman DA, Hayes-Roth F (eds) Pattern-Directed Inference Systems,
  Academic Press, Orlando, pp 313--329

\bibitem[{Howard and Bull(2008)}]{conf/gecco/HowardB08}
Howard GD, Bull L (2008) On the effects of node duplication and
  connection-oriented constructivism in neural {XCSF}. In: Ryan C, Keijzer M
  (eds) Genetic and Evolutionary Computation Conference, {GECCO} 2008,
  Proceedings, Atlanta, {GA}, {USA}, July 12-16, 2008, Companion Material, ACM,
  pp 1977--1984

\bibitem[{Howard et~al(2008)Howard, Bull, and Lanzi}]{conf/gecco/HowardBL08}
Howard GD, Bull L, Lanzi PL (2008) Self-adaptive constructivism in neural {XCS}
  and {XCSF}. In: Ryan C, Keijzer M (eds) Genetic and Evolutionary Computation
  Conference, {GECCO} 2008, Proceedings, Atlanta, {GA}, {USA}, July 12-16,
  2008, ACM, pp 1389--1396

\bibitem[{Howard et~al(2009)Howard, Bull, and Lanzi}]{howard-gecco09}
Howard GD, Bull L, Lanzi PL (2009) Towards continuous actions in continuous
  space and time using self-adaptive constructivism in neural {XCSF}. In: GECCO
  '09: Proceedings of the 11th Annual conference on Genetic and evolutionary
  computation, ACM, New York, NY, USA, pp 1219--1226,
  \doi{http://doi.acm.org/10.1145/1569901.1570065}

\bibitem[{Hurst and Bull(2006)}]{journals/alife/HurstB06}
Hurst J, Bull L (2006) A neural learning classifier system with self-adaptive
  constructivism for mobile robot control. Artificial Life 12(3):353--380

\bibitem[{Hurst et~al(2002)Hurst, Bull, and Melhuish}]{Hurst:2002:TLC}
Hurst J, Bull L, Melhuish C (2002) {TCS} learning classifier system controller
  on a real robot. Lecture Notes in Computer Science 2439:588--600

\bibitem[{Katagami and Yamada(2000)}]{kata-yam}
Katagami D, Yamada S (2000) Interactive classifier system for real robot
  learning. In: IEEE International Workshop on Robot-Human Interaction -
  ROMAN-2000, Osaka, Japan, pp 258--263

\bibitem[{Kohavi and John(1997)}]{KohaviJohn:97}
Kohavi R, John G (1997) Wrappers for feature subset selection. Artificial
  Intelligence 97:273--324

\bibitem[{Koller and Sahami(1996)}]{conf/icml/KollerS96}
Koller D, Sahami M (1996) Toward optimal feature selection. In: International
  Conference on Machine Learning, pp 284--292

\bibitem[{Korkin et~al(1998)Korkin, Nawa, and de~Garis}]{Korkin:1998:SII}
Korkin M, Nawa NE, de~Garis H (1998) A ``spike interval information coding''
  representation for {ATR}'s {CAM-Brain Machine} ({CBM}). Lecture Notes in
  Computer Science 1478:256--??

\bibitem[{Lanzi and Perrucci(1999)}]{lanzi:1999:ERCCPIFMCS}
Lanzi PL, Perrucci A (1999) Extending the representation of classifier
  conditions part {II}: From messy coding to {S}-expressions. In: Banzhaf W,
  Daida J, Eiben AE, Garzon MH, Honavar V, Jakiela M, Smith RE (eds)
  Proceedings of the Genetic and Evolutionary Computation Conference, Morgan
  Kaufmann, Orlando, Florida, USA, vol~1, pp 345--352

\bibitem[{Lanzi and Wilson(2006)}]{lanzi:2006:ch}
Lanzi PL, Wilson SW (2006) Using convex hulls to represent classifier
  conditions. In: GECCO '06: Proceedings of the 8th annual conference on
  Genetic and evolutionary computation, ACM Press, New York, NY, USA, pp
  1481--1488, \doi{http://doi.acm.org/10.1145/1143997.1144240}

\bibitem[{Lanzi et~al(2005{\natexlab{a}})Lanzi, Loiacono, Wilson, and
  Goldberg}]{conf/cec/LanziLWG05a}
Lanzi PL, Loiacono D, Wilson SW, Goldberg DE (2005{\natexlab{a}}) {XCS} with
  computed prediction in continuous multistep environments. In: IEEE Congress
  on Evolutionary Computation, IEEE, pp 2032--2039

\bibitem[{Lanzi et~al(2005{\natexlab{b}})Lanzi, Loiacono, Wilson, and
  Goldberg}]{conf/gecco/LanziLWG05a}
Lanzi PL, Loiacono D, Wilson SW, Goldberg DE (2005{\natexlab{b}}) {XCS} with
  computed prediction in multistep environments. In: Beyer HG, O'Reilly UM
  (eds) Genetic and Evolutionary Computation Conference, {GECCO} 2005,
  Proceedings, Washington {DC}, {USA}, June 25-29, 2005, ACM, pp 1859--1866

\bibitem[{Lanzi et~al(2006)Lanzi, Loiacono, Wilson, and Goldberg}]{lanzi-tile}
Lanzi PL, Loiacono D, Wilson SW, Goldberg DE (2006) Classifier prediction based
  on tile coding. In: Proceedings of the 8th annual conference on Genetic and
  evolutionary computation, ACM, New York, NY, USA, GECCO '06, pp 1497--1504

\bibitem[{Lapicque(1907)}]{lapique}
Lapicque L (1907) Recherches quantitatifs sur l'excitation electrique des nerfs
  traitée comme une polarisation. In: Journal of Physiological Pathology, San
  Francisco, California, USA, vol~9, pp 620--635

\bibitem[{Lin(1991)}]{conf/icml/Lin91}
Lin LJ (1991) Self-improvement based on reinforcement learning, planning and
  teaching. In: Machine Learning, pp 323--327

\bibitem[{Loiacono(2010)}]{loiacono-thesis}
Loiacono D (2010) Rule-based evolutionary systems for generalized reinforcement
  learning. PhD thesis, Politecnico di Milano, Milan, Italy

\bibitem[{Maass(1997)}]{maass}
Maass W (1997) Networks of spiking neurons: the third generation of neural
  network models. Neural networks

\bibitem[{Mahadevan and Connell(1992)}]{Mahadevan/Connell:1992}
Mahadevan S, Connell J (1992) Automatic programming of behavior-based robots
  using reinforcement learning,. Artificial Intelligence 55:311--365

\bibitem[{Michel(2004)}]{webots04}
Michel O (2004) Webots{TM}: Professional mobile robot simulation. International
  Journal of Advanced Robotic Systems 1(1):39--42

\bibitem[{Moioli et~al(2007)Moioli, Vargas, and Zuben}]{conf/iwcls/MoioliVZ07}
Moioli RC, Vargas PA, Zuben FJV (2007) Analysing learning classifier systems in
  reactive and non-reactive robotic tasks. In: Bacardit J, Bernad{\'o}-Mansilla
  E, Butz MV, Kovacs T, Llor{\`a} X, Takadama K (eds) International Workshop on
  Learning Classifier Systems {IWLCS}, Springer, Lecture Notes in Computer
  Science, vol 4998, pp 286--305

\bibitem[{Pipe and Carse(2002)}]{Pipe:2002:FRE}
Pipe AG, Carse B (2002) First results from experiments in fuzzy classifier
  system architectures for mobile robotics. Lecture Notes in Computer Science
  2439:578--587

\bibitem[{Quartz and Sejnowski(1997)}]{QandS}
Quartz SR, Sejnowski TJ (1997) The neural basis of cognitive development: A
  constructivist manifesto. Behavioral and Brain Sciences

\bibitem[{Rocha et~al(2007)Rocha, Cortez, and Neves}]{journals/ijon/RochaCN07}
Rocha M, Cortez P, Neves J (2007) Evolution of neural networks for
  classification and regression. Neurocomputing 70(16-18):2809--2816

\bibitem[{Rumelhart and McClelland(1986)}]{rumelhart86}
Rumelhart D, McClelland J (1986) Parallel Distributed Processing, vol 1 \& 2.
  MIT Press, Cambridge, MA

\bibitem[{Saggie-Wexler et~al(2006)Saggie-Wexler, Keinan, and Ruppin}]{sag-wex}
Saggie-Wexler K, Keinan A, Ruppin E (2006) Neural processing of counting in
  evolved spiking and mcculloch-pitts agents. Artificial Life 12(1):1--16

\bibitem[{Santamaria et~al(1998)Santamaria, Sutton, and Ram}]{santa}
Santamaria JC, Sutton RS, Ram A (1998) Experiments with reinforcement learning
  in problems with continuous state and action spaces. Tech. rep., University
  of Massachusetts

\bibitem[{Schlessinger et~al(2005)Schlessinger, Bentley, and
  Lotto}]{conf/ecal/SchlessingerBL05}
Schlessinger E, Bentley PJ, Lotto RB (2005) Analysing the evolvability of
  neural network agents through structural mutations. In: Capcarr{\`e}re MS,
  Freitas AA, Bentley PJ, Johnson CG, Timmis J (eds) Advances in Artificial
  Life, 8th European Conference, {ECAL} 2005, Canterbury, {UK}, September 5-9,
  2005, Proceedings, Springer, Lecture Notes in Computer Science, vol 3630, pp
  312--321

\bibitem[{Stanley and Miikkulainen(2002)}]{NEAT}
Stanley KO, Miikkulainen R (2002) Evolving neural networks through augmenting
  topologies. Evolutionary Computation 10:99--127

\bibitem[{Stolzmann(1999)}]{stolzmann}
Stolzmann W (1999) Latent learning in khepera robots with anticipatory
  classifier systems. In: Lanzi PL, Stolzmann W, Wilson SW (eds) 2nd
  International Workshop on Learning Classifier Systems, Orlando, Florida, USA,
  pp 290--297

\bibitem[{Studley and Bull(2005)}]{conf/cec/StudleyB05}
Studley M, Bull L (2005) {X}-{TCS}: accuracy-based learning classifier system
  robotics. In: IEEE Congress on Evolutionary Computation, IEEE, pp 2099--2106

\bibitem[{Sutton(1996)}]{sutton-scc}
Sutton RS (1996) Generalization in reinforcement learning: Successful examples
  using sparse coarse coding. In: Advances in Neural Information Processing
  Systems 8, MIT Press, pp 1038--1044

\bibitem[{Sutton and Barto(1998)}]{SaB}
Sutton RS, Barto AG (1998) Reinforcement Learning: An Introduction. MIT Press,
  Cambridge, MA, USA

\bibitem[{Sutton et~al(1999)Sutton, Precup, and Singh}]{sutton-99}
Sutton RS, Precup D, Singh S (1999) Between mdps and semi-mdps: a framework for
  temporal abstraction in reinforcement learning. Artificial Intelligence
  112(1-2):181--211

\bibitem[{Tham(1995)}]{journals/ras/Tham95}
Tham CK (1995) Reinforcement learning of multiple tasks using a hierarchical
  {CMAC} architecture. Robotics and Autonomous Systems 15(4):247--274

\bibitem[{Valenzuela-Rend{\'o}n(1991)}]{Valenzuela-Rendon1991a}
Valenzuela-Rend{\'o}n M (1991) The fuzzy classifier system: a classifier system
  for continuously varying variables. In: Booker LB, Belew RK (eds) Proceedings
  of the 4th International Conference on Genetic Algorithms ({ICGA91}), Morgan
  Kaufmann, pp 346--353

\bibitem[{Watkins(1989)}]{watkins-thesis}
Watkins C (1989) Learning from delayed rewards. PhD thesis, Cambridge
  University, Psychology Dept., Cambridge, UK

\bibitem[{Webb et~al(2003)Webb, Hart, Ross, and Lawson}]{conf/ecal/WebbHRL03}
Webb A, Hart E, Ross P, Lawson A (2003) Controlling a simulated khepera with an
  {XCS} classifier system with memory. In: Banzhaf W, Christaller T, Dittrich
  P, Kim JT, Ziegler J (eds) Advances in Artificial Life, 7th European
  Conference, {ECAL} 2003, Dortmund, Germany, September 14-17, 2003,
  Proceedings, Springer, Lecture Notes in Computer Science, vol 2801, pp
  885--892

\bibitem[{Wilson(1994)}]{Wilson94}
Wilson SW (1994) {ZCS:} {A} zeroth level classifier system. Evolutionary
  {C}omputation 2(1):1--18

\bibitem[{Wilson(1995)}]{wilson:1995:cfba}
Wilson SW (1995) Classifier fitness based on accuracy. Evolutionary
  {C}omputation 3(2):149--175

\bibitem[{Wilson(2000)}]{Wilson1999b}
Wilson SW (2000) Get real! xcs with continuous-valued inputs. In: Learning
  Classifier Systems, From Foundations to Applications, LNAI-1813,
  Springer-Verlag, pp 209--219

\bibitem[{Wilson(2001{\natexlab{a}})}]{Wilson2001a}
Wilson SW (2001{\natexlab{a}}) Function approximation with a classifier system.
  In: Spector L, Goodman ED, Wu A, Langdon WB, Voigt HM, Gen M, Sen S, Dorigo
  M, Pezeshk S, Garzon MH, Burke E (eds) Proceedings of the Genetic and
  Evolutionary Computation Conference (GECCO-2001), Morgan Kaufmann, San
  Francisco, California, USA, pp 974--981

\bibitem[{Wilson(2001{\natexlab{b}})}]{Wilson2001b}
Wilson SW (2001{\natexlab{b}}) Mining oblique data with {XCS}. In: Lanzi PL,
  Stolzmann W, Wilson SW (eds) Advances in learning classifier systems, third
  international workshop, {IWLCS} 2000, LNCS, vol 1996, Springer, pp 158--176

\end{thebibliography}

\end{document}